\documentclass[review]{elsarticle}
\usepackage{amsmath,amsfonts,amssymb}
\usepackage{graphicx}
\usepackage{bm}
\usepackage{subfig}
\usepackage{multirow}
\usepackage{makecell}
\usepackage{url}
\usepackage{booktabs}
\usepackage{soul}
\usepackage{placeins}
\usepackage{floatrow}
\newfloatcommand{capbtabbox}{table}[][\FBwidth]
\usepackage[colorlinks=true, allcolors=blue]{hyperref}
\usepackage{lineno,hyperref}
\modulolinenumbers[5]

\journal{Journal of \LaTeX\ Templates}

\begin{document}

\begin{frontmatter}

\title{SSPNet: Scale and spatial priors guided generalizable and interpretable pedestrian attribute recognition}
\author[a]{Jifeng~Shen\corref{mycorrespondingauthor}}
\cortext[mycorrespondingauthor]{Corresponding author}
\ead{shenjifeng@ujs.edu.cn}

\author[a]{Teng~Guo}
\author[b]{Xin~Zuo}
\author[c]{Heng~Fan}
\author[d]{Wankou~Yang}
\address[a]{School of Electrical and Information Engineering, Jiangsu University, Zhenjiang, 212013, China}
\address[b]{School of Computer Science and Engineering, Jiangsu University of Science and Technology, Zhenjiang, 212003, China}
\address[c]{Department of Computer Science and Engineering, University of North Texas, Denton, TX 76207, USA}
\address[d]{School of Automation, Southeast University, Nanjing, 210096, China}

\begin{abstract}
Global feature based Pedestrian Attribute Recognition (PAR) models are often poorly localized when using Grad-CAM for attribute response analysis, which has a significant impact on the interpretability, generalizability and performance. Previous researches have attempted to improve generalization and interpretation through meticulous model design, yet they often have neglected or underutilized effective prior information crucial for PAR. To this end, a novel Scale and Spatial Priors Guided Network (SSPNet) is proposed for PAR, which is mainly composed of the Adaptive Feature Scale Selection (AFSS) and Prior Location Extraction (PLE) modules. The AFSS module learns to provide reasonable scale prior information for different attribute groups, allowing the model to focus on different levels of feature maps with varying semantic granularity. The PLE module reveals potential attribute spatial prior information, which avoids unnecessary attention on irrelevant areas and lowers the risk of model over-fitting. More specifically, the scale prior in AFSS is adaptively learned from different layers of feature pyramid with maximum accuracy, while the spatial priors in PLE can be revealed from part feature with different granularity (such as image blocks, human pose keypoint and sparse sampling points). Besides, a novel IoU based attribute localization metric is proposed for Weakly-supervised Pedestrian Attribute Localization (WPAL) based on the improved Grad-CAM for attribute response mask. The experimental results on the intra-dataset and cross-dataset evaluations demonstrate the effectiveness of our proposed method in terms of mean accuracy (mA). Furthermore, it also achieves superior performance on the PCS dataset for attribute localization in terms of IoU. 
Code will be released at https://github.com/guotengg/SSPNet.
\end{abstract}

\end{frontmatter}
%\linenumbers

\section{Introduction}
\label{sec:intro}
Pedestrian Attribute Recognition (PAR) focuses on identifying the physical characteristics of pedestrians, such as gender, clothing and hairstyle from images or videos captured in various environments, which is a challenging problem in computer vision. 
The PAR model can be used to serve a wide range of downstream tasks, such as video surveillance systems, intelligent transportation system and intelligent robots.
Recently, significant progress has been made in PAR as a result of advancements in deep learning, which can also contribute to closely related topics such as pedestrian retrieval \cite{li2018richly} and pedestrian re-identification \cite{yang2019towards}.

\begin{figure}[h]
    \centering
    \includegraphics[width=.8\textwidth]{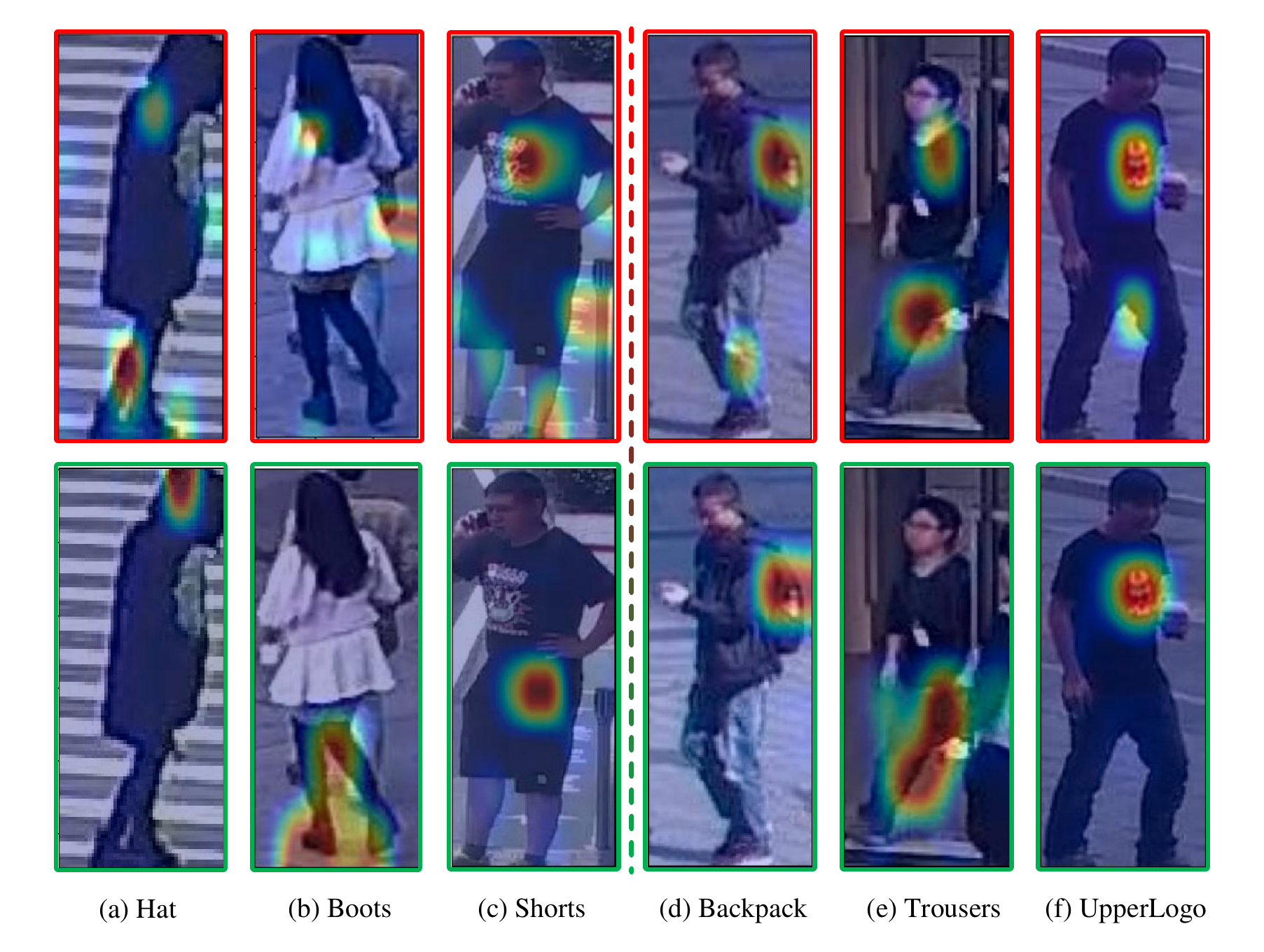}
    \caption{The visualization of previous PAR model (red border in the first row) and our proposed model (green border in the second row) for the corresponding attributes. ((a), (b) and (c) with red border show wrong response locations. (d), (e), (f) with red border show the coexistence of correct response position and irrelevant position. The images with the green border are the effect of our proposed method after solving the corresponding problem.)}
    \label{defect}
\end{figure}

Previous studies typically solve the attribute recognition problem through multi-task learning~\cite{sudowe2015person} and multi-label learning~\cite{li2020feature}. They focus on discriminative features whose positions are learned implicitly. However, as illustrated in Fig.~\ref{defect}, several attributes response heatmap (visualized by Grad-CAM~\cite{selvaraju2017grad}) learned from these SOTA  methods~\cite{yu2016weakly,liu2017hydraplus,sarafianos2018deep}  are unreasonable and perplexing.
For example, in Fig.~1(a), Fig.~1(b) and Fig.~1(c) with red border, the attribute response heatmaps for the ``Hat'', ``Boots'' and ``Backpack'' attributes indicate a focus on incorrect positions, which are uninterpretable and also counter-intuitive for human beings. 
We argue that one plausible explanation for this phenomenon is the different learning needs of attributes with various granularities. The failure to accommodate these diverse learning needs when using feature maps presents a challenge in identifying discriminative features on unsuitable maps. For instance, the model struggles to learn discriminative features from low-resolution feature maps when dealing with fine-grained attributes, while coarse-grained attributes may suffer from background noise introduced by high-resolution feature maps. Consequently, this mismatch in feature maps leads to heatmap responses being in incorrect positions.
In Fig.~1(d), Fig.~1(e) and Fig.~1(f) with red border, the heatmaps of ``Shorts'', ``Trousers'' and ``UpperLogo'' give response to both the correct and  unrelated locations. These irrelevant regions contain features that stem from dataset biases, which introduce interference and hinder the model's generalization performance.
We argue that the absence of spatial prior guidance during the model's learning process is the primary cause of this phenomenon. For instance, in the absence of spatial priors, the model is compelled to search for the discriminative features of a hat across the entire feature map, making it susceptible to the influence of irrelevant areas. However, with the aid of spatial priors, which provide partial features near the head, the task of locating discriminative features becomes notably more straightforward. Therefore, the lack of spatial prior guidance makes model more susceptible to dataset bias, resulting in poor generalization performance.
To this end, we aim to answer the following two questions in order to address these issues.

\textbf{How to deal with different learning needs for attributes with various granularity~?} One solution is to provide multi-scale feature maps for attributes with varying granularity, facilitating the extraction of discriminative features.

Previous works~\cite{zhong2021improving} introduce feature pyramid structure to obtain a multi-scale feature map, which contains discriminative features with different granularity.
However, they often employ a fixed feature map scale for all attributes, which fails to fully exploit the sensitivity of features related to attribute granularity. In this paper, we introduce an Adaptive Feature Scale Selection (AFSS) module that dynamically selects an appropriate feature map based on the mean accuracy (mA) metric during model training. It effectively addresses the challenge of recognizing attributes with varying granularities, enhancing model interpretability and achieving higher IoU values in WPAL. In our experiments, it is interesting to find that the AFSS module selects high-resolution feature maps for fine-grained attributes (such as ``Hat'' and ``Boots''), while favoring low-resolution feature maps for coarse-grained attributes (like ``Age'' and ``Gender'')

\textbf{How to get more reasonable attribute locations with spatial prior guidance~?} 
Providing spatial prior knowledge is typically necessary to assist the model in avoiding poor generalization by neglecting irrelevant regions, as well as to assist the model in focusing on discriminative features in the correct attribute regions.

Existing studies on such spatial knowledge have focused on developing learnable modules~\cite{liu2019localization} for acquiring attribute locations or spatially transforming regions based on human posture~\cite{li2018pose} to facilitate pedestrian attribute recognition. However, most of the human attributes (e.g. ``glasses'', ``shirt styles'' and ``shoes'') have relatively stable locations in human body and have strong spatial characteristics. This significant spatial characteristic is known as a spatial prior. It acts as a spatial constraint, guiding the model to concentrate on the accurate attribute locations. This helps in mitigating learning biases from other regions, ultimately enhancing the model's generalization performance. In this paper, these spatial prior knowledge can be utilized directly without the help of  complicated extra model. Beside, we use learnable offset points based on such spatial prior to balance the constraint of prior knowledge and the freedom of model learning.

Based on these ideas, we propose a SSPNet to learn generalizable and interpretable PAR model, which mainly comprise of AFSS and PLE modules based on the multi-label classification framework. The AFSS module can select feature map with proper scales for attributes of various granularity, while the PLE module aims to incorporate the spatial prior knowledge of attributes into the model learning process. From another perspective, AFSS identifies the most suitable scale of feature map, while PLE utilizes spatial priors to find the appropriate location of attribute in the feature map. Besides, both the AFSS and PLE modules can be easily applied after different backbone in PAR model.

To summarize, our main contributions are fourfold:
\begin{itemize}
	\item An AFSS module is proposed to select feature map with proper scales for attributes of varying granularity, leading to improved performance and better interpretability.
	\item A PLE module is proposed to assist PAR by guiding the model to learn the correct attribute location and discriminative features for better generalization.
	\item A weakly supervised attribute localization with IoU values are achieved based on the attribute response heatmaps, showing better interpretability.
	\item The proposed method has achieved state-of-the-art results on PA100k dataset, as well as better generalization performance in cross-dataset tests. 
\end{itemize}

The rest of this paper is organized as follows. Section 2 introduces the related work published in recent years. Section 3 describes our proposed method in detail. The experimental results are given in Section 4 and we conclude the paper in Section 5.

\section{Related work}
\subsection{Pedestrian attribute recognition}
Deep learning-based methods have achieved remarkable progress in PAR. Convolutional Neural Networks (CNNs) have been widely adopted for feature extraction and classification due to their excellent ability to automatically learn discriminative features. 
Jia et al.~\cite{jia2021rethinking} adopt a classical approach by leveraging global features to achieve PAR. In order to make the model focused on the attributes, Wu et al.~\cite{wu2020distraction} introduce attention mechanisms and Yang at al.~\cite{yang2016attribute} propose a method that utilizes part features to eliminate interference from irrelevant regions.
Furthermore, Wang et al.~\cite{wang2017attribute} and Tang et al.~\cite{tan2020relation} employ the internal relationships between attributes as a way to guide model learning.

\textbf{Scale knowledge.} Pedestrians exhibit attributes at various scales in natural images. The inclusion of multi-scale feature (eg. feature pyramid network~\cite{lin2017feature}) aids the model in capturing a wide range of feature information across different scales. Consequently, methods~\cite{li2020feature,sarafianos2018deep} that leverage multi-scale feature demonstrate enhanced accuracy and robustness in PAR. Yang et al.~\cite{yang2021cascaded} use multi-scale features for cascade and split learning in PAR. Zhong et al.~\cite{zhong2021improving} propose a multi-scale spatial calibration module to gather information and build long-range dependencies among pyramid feature maps.
In this paper, compared to PAR methods that utilize feature pyramid network, our approach incorporates scale prior to select appropriate scale of feature map for each attribute group adaptively.

\textbf{Spatial prior.} In addition to using the attribute annotation information, more spatial prior information can be adopted from pedestrian images and videos to assist with PAR.
For instance, methods leveraging attribute positions~\cite{li2017learning} and pedestrian body posture~\cite{li2019pedestrian} as learnable guides have demonstrated their utility in PAR. The PGDM proposed by Li et al.~\cite{li2018pose} incorporates pedestrian body structure so that the part-based results are fused with global body-based results for improved attribute prediction. Lu et al.~\cite{lu2023orientation} and Fan et al.~\cite{fan2023parformer} employ the human orientation as an aid in PAR.
In this paper, our method introduces a new feature learning mechanism rooted in spatial prior, which can strike a balance between the constraints imposed by prior knowledge and the flexibility of model learning.

\subsection{Weakly supervised attribute localization}
The weakly supervised localization with image-level supervision is an important problem, which greatly reduces the cost of annotations. 
In this area, weakly supervised attribute localization is conceptually similar to weakly supervised object detection. 
The former one focuses on fine-grained objects or abstract concepts, such as parts of a pedestrian (hat or shoes) and age, whereas the latter one focuses on the whole object, such as a pedestrian's location.

Previous studies usually implement weakly supervised localization based on CAM~\cite{zhou2016learning} and Grad-CAM, which gives a good inspiration for attribute localization. As a result, weakly supervised attribute localization has followed in a similar way.
Wang et al.~\cite{wang2013weakly} propose a weakly supervised method for learning scene parts and attributes from a collection of images associated with attributes in text.
Yu et al.~\cite{yu2016weakly} use novelly designed detection layers to discover mid-level attribute features and the rough shapes of pedestrian attributes are inferred by performing clustering on a fusion of activation maps. 
The ALM~\cite{tang2019improving} learns regional features for each attribute at multiple levels and adaptively discovers the most discriminative regions for attribute recognition and localization.
Weng et al.~\cite{weng2023exploring} employ attention mechanisms to achieve attribute localization, concurrently aiding in the task of PAR.

In this paper, our method employs improved Grad-CAM (called Grad-CAM-P) method in weakly supervised attribute localization task to output a set of bounding boxes. Beside, we propose a localization method based on bounding box and also provide a quantitative metric (IoU) for weakly supervised pedestrian attribute localization in the first time.

\section{Method}
\subsection{Problem definition}
PAR is commonly formulated as a multi-label classification problem.
Given an attribute dataset with $N$ pedestrian samples $D = \left\{\left(X_{i}, Y_{i}\right)\right\}_{i=1}^{N}$, $X_{i} \in \mathbb{R}^{W\times H\times 3}$ and $Y_{i} \in\{0,1\}^{M}$ denote the $i^{th}$ pedestrian image and its attribute label respectively. $M$ represents the number of attributes, $W$ and $H$ are the width and height of image. 
In PAR task, given an input $X_{i}$, and the PAR model $f_{{A}}$ outputs a prediction vector $\bm{a}$, which is formulated in Eq.~\ref{eq:PAR}.
\begin{equation}
\begin{aligned}
\bm{a} = f_{{A}}(\mathcal{F}(\Psi_{b}({X}; \bm{\theta}_{{A}}); \bm{\theta}_{{S}}); \bm{\theta}_{{P}})
\end{aligned}
\label{eq:PAR}
\end{equation}

\noindent where $\Psi_{b}(\cdot)$ is a feature extraction function (eg. ResNet50~\cite{he2016deep}, Swin Transformer~\cite{liu2021swin}, etc.) with parameter $\bm{\theta}_{{A}}$. The $\mathcal{F}$ is the AFSS module with scale prior $\bm{\theta}_{{S}}$, while the $\bm{\theta}_{{P}}$ is the spatial prior used in PLE module. $\bm{a} \in \mathbb{R}_{[0,1]}^M$ is the attribute prediction.

Weakly-supervised Pedestrian Attribute Localization (WPAL) aims at locating of object parts or abstract concepts based on image-level attribute label, without bounding box annotations. 
Given an image $X$ and attribute type $c$, we aim to obtain a set of bounding boxes $\bm{b}$ using function $f_{orm}(\cdot)$ as shown in Eq.~\ref{eq:CAM2}. 
\begin{equation}
\begin{aligned}
A_{c}(X) = f_{cam}(\mathcal{F}(\Psi_{b}(X)),c)  %\footnotemark[1] \
\end{aligned}
\label{eq:CAM}
\end{equation}
\begin{equation}
\begin{aligned}
\bm{b} = f_{orm}(A_{c}(X))
\end{aligned}
\label{eq:CAM2}
\end{equation}

\noindent where $A_{c}(.)$\footnote{Here, we omit the model parameters $\bm{\theta}_{{A}}$  and $\bm{\theta}_{{S}}$ in Eq.~\ref{eq:CAM}
for clarity.} is the response heatmap of Grad-CAM-P function $f_{cam}(.)$ for the $c^{th}$ attribute, and $f_{orm}(.)$ is a outer rectangle function to enclose binary masks generated from heatmap.

\begin{figure}[h]
	\includegraphics[width=1\linewidth]{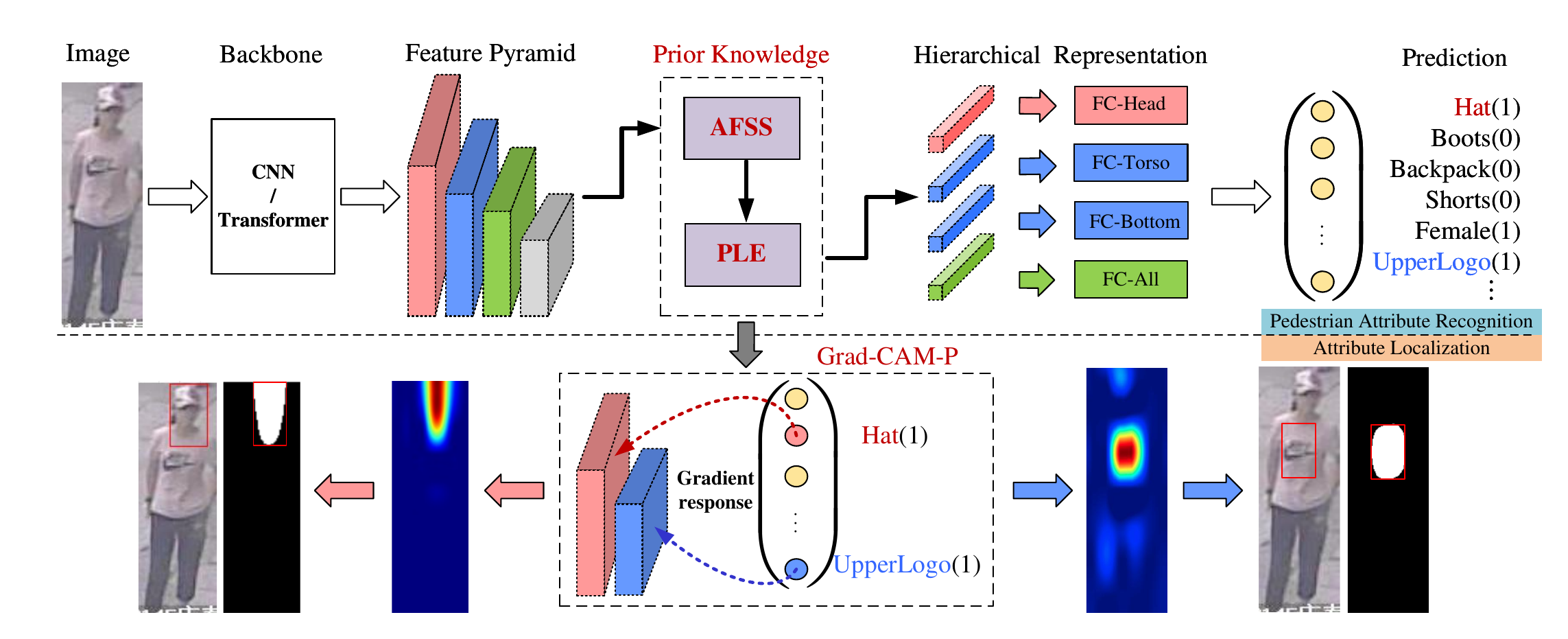}
	\caption{Overview of the proposed pipeline. (The workflow of the proposed PAR method is shown in the upper part, while the proposed weakly supervised attribute localization method is shown in the bottom part.)}
\label{fig:overview}
\end{figure}

\subsection{Proposed method}

\subsubsection{Overview} 
Our proposed PAR model is based on the multi-label classification framework, which mainly comprises of Adaptive Feature Scale Selection (AFSS) module and Prior Location Extraction (PLE) module. As shown in column ``Groups'' and ``Attributes'' of Tab.~\ref{tab:attribute information}, all pedestrian attributes have been divided into four groups based on spatial prior knowledge. An overview of the proposed pipeline is shown in Fig.~\ref{fig:overview}, where a feature pyramid is firstly obtained by extracting the multi-scale features from the input image through the backbone. 
Following that, the feature maps selected by the AFSS module are fed into the PLE module where we investigate three different strategies for leveraging the spatial priors and feature maps.
Subsequently, the feature vectors obtained from the PLE module are aggregated and used as input for the hierarchical representation module, which produces the attribute recognition results.
Furthermore, after obtaining the PAR model, we employ Grad-CAM-P to generate attribute heatmaps, which can be utilized to output bounding box for the each attribute.
\begin{figure}[h]
	\includegraphics[width=.9\linewidth]{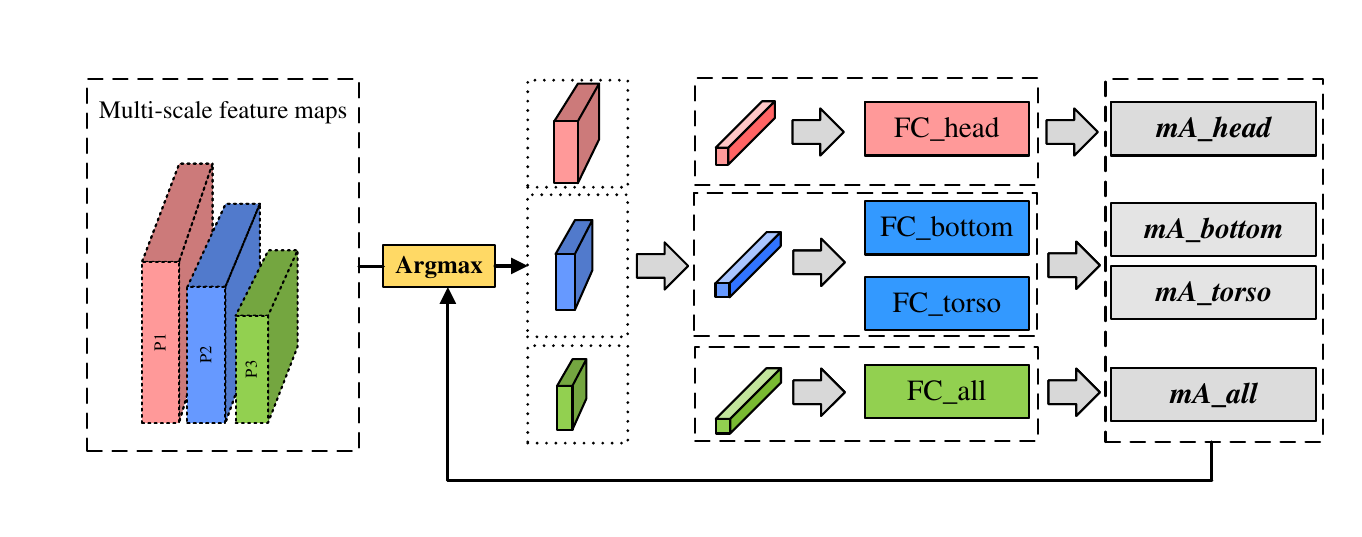}
	\caption{Schematic workflow of the AFSS module. (This module selects the feature map that maximizes the mA during the model training.)}
	\label{fig:AFSS}
\end{figure}

\subsubsection{Adaptive feature scale selection (AFSS)}
The AFSS module adaptively selects the appropriate scale of feature map in the feature pyramid for different attribute groups, based on the best classification result in terms of mean accuracy (mA). 
The workflow of the AFSS module is illustrated in Fig.~\ref{fig:AFSS}.
Given an input of pedestrian image, we obtain three feature maps from the classical feature pyramid network, which are denoted as $P_1$, $P_2$, and $P_3$ with down-sampling ratio of $1/4$,  $1/8$ and  $1/16$ respectively.
Then, the AFSS module evaluates the attribute recognition performance achieved from feature maps P1, P2, and P3 using the mA scores. In the model training phase, the AFSS module iteratively updates the choices of optimal scale based on the maximum mA value for each attribute group. Subsequently, in the model testing phase, the AFSS module uses the learned optimal feature map for each attribute group. The primary advantage of the AFSS module is its dynamic selection of the optimal feature map based on the model's performance feedback, seamlessly integrated into the training process.
Eq.~\ref{eq:AFSS} shows the selection of optimal feature map $x^{*}$ in the training phase: 
\begin{equation}
    x^* = \arg\max_{x} (mA(f_{A}(x),Y)), {x} \in \{P_1,P_2,P_3\}
\label{eq:AFSS}
\end{equation}
where $mA(\cdot)$ is the mean accuracy function, $f_{A}(x)$ represents the attributes prediction result with feature map $x$.

We find that attributes in the same group are closely related with similar size and location. For example, the attributes listed in the ``Head'' group are related to the parts above the shoulders of a person. Since the human head takes up a smaller proportion of the human body,  the head-related attributes  also give response in a smaller area size. Besides, attributes that may related to the whole human body are divided into the ``All'' group. These attributes are commonly more abstract concepts, which can be determined by observing the full human body. Intuitively, the same group attributes can share the same scale of feature map, and this idea is supported by Tab.~\ref{tab:AFSS role} in section.~\ref{sec:Ablation Study}.

\begin{figure*}[h]
	\centering
	\captionsetup{aboveskip=0pt}
	\captionsetup{belowskip=0pt}
	\subfloat[PLE-R]{	
		\includegraphics[width=0.313\linewidth]{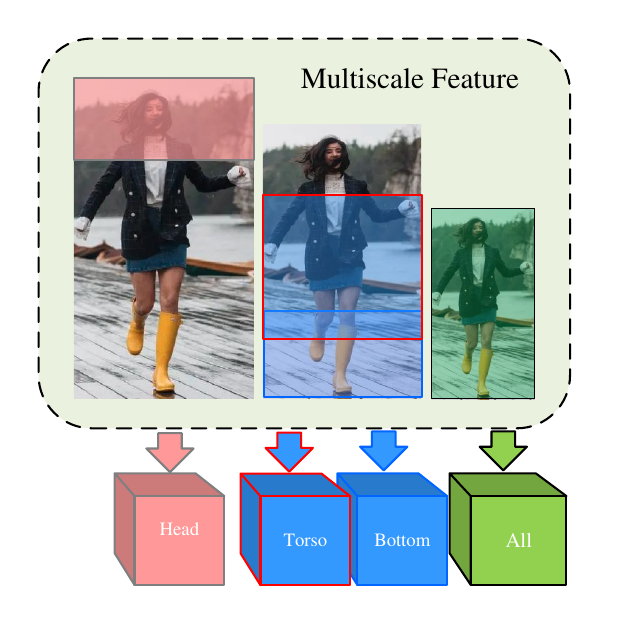}
		\label{fig:PLE-R}
	}\hfill
	\subfloat[PLE-K]{
		\includegraphics[width=0.313\linewidth]{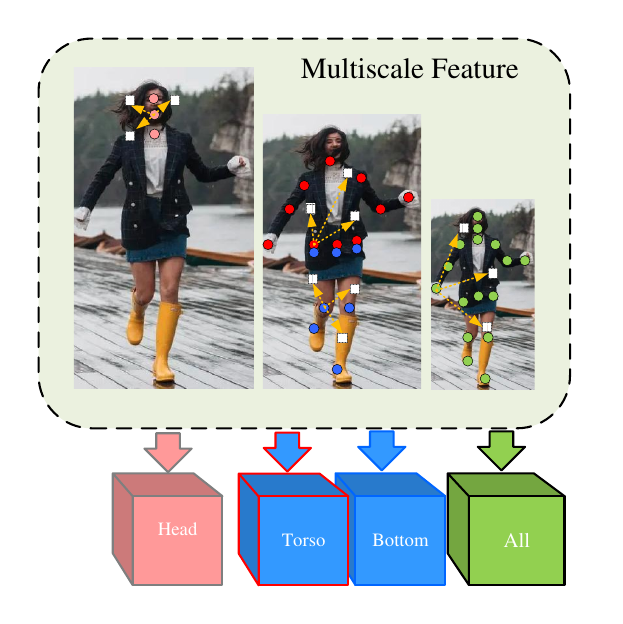}
		\label{fig:PLE-K}
	}\hfill
	\subfloat[PLE-S]{
		\includegraphics[width=0.313\linewidth]{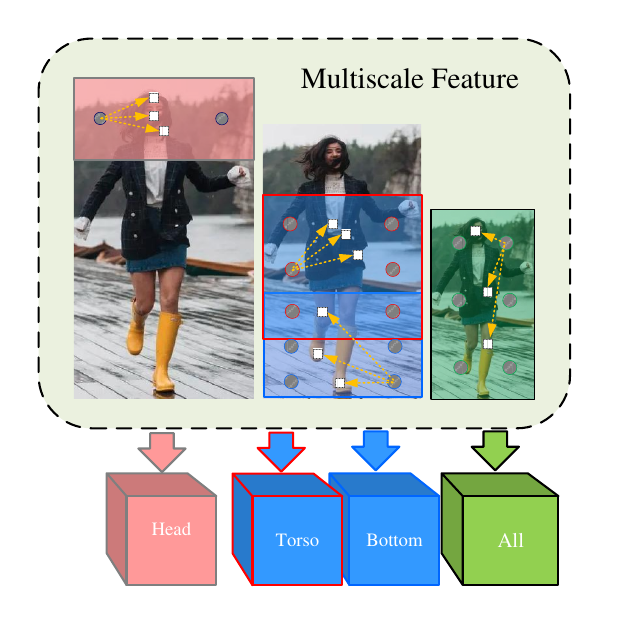}
		\label{fig:PLE-S}
	}
	\caption{PLE modules with three different spatial priors. (a) PLE-R (The rectangles with colors are prior regions designed on the image height), (b) PLE-K (The circles with colors are the human keypoints), and (c) PLE-S (The circles are the sparse sampling points for prior regions). Small white squares indicate learnable offset points. Cubes of different colors represent feature vectors for different attribute groups.}
	\label{fig:PLE}
\end{figure*}

\subsubsection{Prior location extraction (PLE)}
The PLE module aims to incorporate the spatial priors of attributes into the model learning process, which avoids unnecessary attention on irrelevant areas and also lowers the risk of model over-fitting. In this section, three different spatial priors are investigated which will be detailed in the following. 

\textbf{Region-based PLE (PLE-R).}
In this method, the feature maps are divided into several sub-regions along the vertical direction to take advantage of existing knowledge of the regions where attributes may exist.
We perform an overlay and averaging operation on the pedestrian images in the dataset, resulting in a prominent outline of the human body.
Motivated by this, the relatively fixed position of human parts can be used as an important hint during the attribute feature learning process. More specifically, we can crop image to get prior regions for pedestrian attributes.
In our implementation, the unrelated feature areas are cropped out, and only the features in the prior regions are kept as input features to the later stage for attribute recognition. The overlaid and averaged pedestrian images are showcased in Fig.~\ref{fig:add image}.

We crop feature map into blocks based on height designed for the four attribute groups and finally get four rectangular areas with colors as shown in Fig.~\ref{fig:PLE-R}.  
The feature vector of each attribute $F_{g}$ in each group is formulated in Eq.~\ref{eq:PLE-R}.
\begin{equation}
\begin{aligned}
     F_{g} = \Psi_{agr}(\Phi_{reg}(x_g^*,R_{g})), g \in \{All, Head, Torso,Bottom\}
\end{aligned}
\label{eq:PLE-R}
\end{equation}

\noindent where $\Phi_{reg}(.)$ is a feature extraction function which outputs feature from prior region $R_{g}$. $x^*_g$ is the optimal feature of attribute group $g$. $\Psi_{agr}(.)$ is the feature aggregation function which outputs the feature vector $ F_{g}$ for attribute group~$g$. In this paper, feature pyramid pooling method can be utilized as $\Psi_{agr}(.)$ to obtain discriminative features for each attribute group. 
The specific height data of the feature maps cropped for each attribute group is displayed in Tab.~\ref{tab:attribute information}.

\textbf{Human Keypoints-based PLE (PLE-K).}
Region-based PLE can be considered as a set of densely sampled points in a region, which usually includes background information irrelevant to the attributes.
To leverage finer spatial priors, human keypoints are an excellent choice as they are closely associated with the regions relevant to attributes. For example, keypoints at the head are closely related to region associated with attributes ``Hat'' and ``Glasses''.

We begin by inputting the original images into the pre-trained SimCC~\cite{li2022simcc} model to obtain 16 human keypoints for each pedestrian.
As shown in different coloured points on the body of Fig.~\ref{fig:PLE-K}, these human keypoints are simply mapped to the different feature maps determined by the AFSS module. 
Next, each keypoint is utilized as a reference point, and  keypoints with the same colour represent in the same attribute group. Besides, each keypoint is associated with $M$ learnable offset points, as formulated in Eq.~\ref{eq:offset-K}. These offset points are shown as small white squares in Fig.~\ref{fig:PLE-K}.

\begin{align}
\begin{aligned}
\Phi_{kpt}(x_g^*, P_{n}^{K}) = \sum_{m=1}^{M} A_{m} \cdot W_m \cdot x_g^*\left(P_{n}^{K}+\Delta P_{n}^{K} \right), n \in \{1,2,...,16\}
\end{aligned}
\label{eq:offset-K}
\end{align} 
\noindent where $\Phi_{kpt}$ is a keypoint based feature extraction function. $A_{m}$ is the learnable weight parameter and $W_{m}$ is the $m^{th}$ offset weight. $P_{n}^{K}$ denotes the $n^{th}$ human keypoint, while $\Delta P_{n}^{K}$ is the corresponding 2D offset vector for the $P_{n}^{K}$. Different from the deformable convolution~\cite{Dai_2017_ICCV}, Eq.~\ref{eq:offset-K} mainly focuses on each individual human keypoint. The specific keypoint indexes selected for attribute groups are provided in Fig.~\ref{fig:keypoint range}.

The offset vectors are responsible for collecting discriminative features that are relevant to the corresponding attribute group. These features are then aggregated using the function $\Psi_{agr}(.)$ to generate the attribute group-specific feature vector $F_{g}$, denoted by the four colorful cubes in Fig.~\ref{fig:PLE-K}, and is formulated in Eq.~\ref{eq:PLE-K}.
\begin{align}
\begin{aligned}
         F_{g} = \Psi_{agr}(\Phi_{kpt}(x_g^*,P_{n}^{K}))
\end{aligned}
\label{eq:PLE-K}
\end{align} 

\textbf{Sparse sampling points-based PLE (PLE-S).}
Although keypoints based feature extraction function can obtain attribute-related features, it relies on additional pose estimation modules which requires additional computation and also brings unavoidable keypoints location errors.
To remedy this, a natural solution is sparse sampling on the attributes related regions, which is also simple and easy to implement. 

As shown in Fig.~\ref{fig:PLE-S}, our proposed approach samples points from each prior region uniformly. The sparse sampling positions are fixed with a sampling ratio (discussed in Section.~\ref{sec:Ablation Study}). 
Then, similar to PLE-K,  we also learn $M$ offset points based on each sparse point through Eq.~\ref{eq:offset-S} and aggregate features from sparse sampling points by Eq.~\ref{eq:PLE-S}. 
\begin{equation}
    \Phi_{sps}(x_{g}^*, P_{n}^{S}) = \sum_{m=1}^{M} A_{m} \cdot W_m \cdot{x_{g}^*}\left(P_{n}^{S}+\Delta P_{n}^{S} \right), n \in \{1,2,...,N\}
\label{eq:offset-S}
\end{equation}

\begin{equation}
	F_{j} = \Psi_{agr}(\Phi_{sps}(x_g^*, P_{n}^{S}))
\label{eq:PLE-S}
\end{equation}

\noindent where $\Phi_{sps}(.)$ denotes sparse sampling based feature extraction function, and $P_{n}^{S}$ is the $n^{th}$ sparse sampling point. $\Delta P_{n}^{S}$ means a 2D offset vector for $P_{n}^{S}$. 
%Eq.~\ref{eq:PLE-S} mainly focuses on each individual sparse sampling point compared to the deformable convolution.

PLE-R does not learn additional information from spatial priors, such as learnable offset points, which can be considered as a baseline method. Additionally, PLE-R employs the coarsest spatial prior, which involves image blocks. PLE-K and PLE-S both require the use of learnable offset points.  PLE-K utilizes the finest spatial prior, but requires additional computations to obtain each pedestrian's keypoints. PLE-S adopts a balanced approach, which sparsely samples the image and leverages learnable offset points to capture more useful features.

\subsubsection{Hierarchical representation}

In this section, we introduce the hierarchical attribute recognition and hierarchical loss function that works in conjunction with the AFSS and PLE modules. Initially, the AFSS selects suitable feature map for the PLE adaptively which outputs feature vectors for each group of attributes. Then, feature vectors are fed into corresponding fully connected layers designed for each group of attributes. As shown in Fig.~\ref{fig:overview}, the four FC layers are independent with each other for the loss computation during the training. We refer to these fully connected layers and losses of each group as the hierarchical attribute recognition and hierarchical loss function respectively.

\textbf{Hierarchical attribute recognition.}
In the previous section, we discussed the division of attributes into four groups, namely ``All," ``Head," ``Torso," and ``Bottom." As illustrated in Fig.~\ref{fig:overview}, for attribute recognition within each group, we utilize feature vectors extracted from the corresponding prior region or points as input to a dedicated FC layer. This approach ensures that only relevant feature vectors from the corresponding prior regions are considered, while excluding irrelevant feature vectors from other regions. The hierarchical attribute recognition process involves a total of four FC layers, with each FC layer corresponding to an attribute group.

\textbf{Hierarchical loss.}
The PAR task can be considered as multiple binary classification tasks in this paper. The output of the final layer is processed by a sigmoid function to obtain the prediction of each attribute. 
Therefore, the loss function of our method can be defined as a weighted binary cross-entropy loss~\cite{jia2021rethinking} (Eq.~\ref{eq:loss}):
\begin{equation}
    Loss^{bce} = \frac{1}{N} \sum_{i = 1}^{N} \sum_{j = 1}^{M} \omega_{j}\left(Y_{i j} \log \left(\sigma\left(\bm{a}_{i j}\right)\right)+\left(1-Y_{i j}\right) \log \left(1-\sigma\left(\bm{a}_{i j}\right)\right)\right)
\label{eq:loss}
\end{equation}

\noindent where $\bm{a}_{ij}$ is the output of the model for the $j^{th}$ attribute of the $i^{th}$ pedestrian picture, $\sigma(\cdot)$ is a sigmoid function and $\omega_j$ is an imbalance weight (Eq.~\ref{eq:imb}), which is calculated based on the number of positive and negative samples of the $j^{th}$ attribute to alleviate the imbalance of positive and negative sample distribution.
\begin{equation}
    \begin{array}{c}
w_{j} = \left\{\begin{array}{ll}
e^{1-r_{j}}, & y_{i j}<0 \\
e^{r_{j}}, & y_{i j} \geq 0
\end{array}\right.
\end{array}
\label{eq:imb}
\end{equation}
where $r_{j}$ is the proportion of positive samples of the $j^{th}$ attribute in the training set.
Finally, hierarchical Loss can be defined in following Eq.~\ref{eq:total loss}:
\begin{equation}
\begin{aligned}
   Loss_{total} = \sum_{g}Loss^{bce}_{g}
\end{aligned}
\label{eq:total loss}
\end{equation}

\noindent where $Loss^{bce}_{g}$ and $Loss_{total}$ represent the loss in the attribute group $g$ and the total loss for all groups respectively.

\subsubsection{Weakly supervised attribute localization}

A improved version of Grad-CAM is employed for weakly supervised attribute localization, dubbed as Grad-CAM-P (Grad-CAM for pixel). It provides pixel-level feature gradient responses instead of the channel-level feature gradient responses obtained by the original Grad-CAM. The computation of Grad-CAM-P is shown as Eq.~\ref{eq:Grad-CAM-P}:
\begin{equation}
\begin{aligned}
L_{{Grad-CAM-P}}&=\operatorname{ReLU}\left(\sum_{k}{\alpha_k^j} A^{k}\right) \\
{\alpha_k^j}&=\frac{\partial{\bm{a}}_j}{\partial A_{x,y}^{k}}
\end{aligned}
\label{eq:Grad-CAM-P}
\end{equation}

\noindent where $A^{k}$ represents $k^{th}$ channel of feature map $A$. $\bm{a}_{j}$ means model prediction result for $j^{th}$ attribute. 
${\alpha_k^j}$ represents the response gradient of $A ^ k$ for the $j^{th}$ attribute.
$A _{x,y} ^k$ means feature value at position coordinates $(x, y)$ of $A^{k}$. 

Unlike multi-class classification, the feature maps obtained from the feature extraction layer contain feature information for multiple categories to ensure the recognition of all existing labels in multi-label classification tasks. Then, the non-zero channel-level gradient response will weightedly sum the features of other category in the feature map, whose pixel-level gradient response should be smaller or even zero, which results in the final computed heatmap containing responses of other category. To address this issue, we use pixel-level response values to weight the feature maps and generate response heatmaps, which can suppress the appearance of irrelevant regions.

Finally, we use feature map selected by AFSS and Grad-CAM-P to generate attribute response heatmap. After binarizing the  heatmap, we convert the foreground regions into one bounding box for attribute localization with simple external rectangle operation.

\section{Experiments}
Our model is implemented with PyTorch 1.9.1 and all the experiments are conducted on Ubuntu 20.04 with NVIDIA GeForce RTX 3080Ti GPU, 32G Memory and 3.70 GHz Intel Core i9-10900K CPU. Pedestrian images are resized to 256$\times$192 as inputs. The training takes 30 epochs with a batch size of 32. The model parameters are updated using the Adam optimizer with initial learning rate of 0.0001. The Strong Baseline~\cite{jia2021rethinking} is used as our baseline method for the ablation experiments.
\subsection{Dataset and metrics}
The PA100K, RAP v1 and UPAR datasets are used for the experiments in this paper which are all public available.

\textbf{PA100k dataset~\cite{liu2017hydraplus}.} The PA100k dataset is constructed by images captured from 598 real outdoor surveillance cameras, containing a total of 100,000 pedestrian images. Each image is annotated with 26 attributes~\cite{wang2022pedestrian}. Following~\cite{liu2017hydraplus,su2017pose}, the dataset is randomly divided into 80,000, 10,000 and 10,000 images for training, validation and testing respectively.

\textbf{RAP v1 dataset~\cite{li2018richly}.} It is a large-scale person re-identification dataset that focuses on predicting various attributes such as gender, age and clothing etc. This dataset comprises 41,585 pedestrian images gathered from 26 surveillance cameras, encompassing 69 binary attributes. 
%Existing methods primarily emphasize 51 attributes, each with a proportion exceeding 1$\%$. 
The training set of this dataset comprises 33,268 images, with the remaining images is used for testing.

\textbf{PETA dataset.}~\cite{deng2014pedestrian}
It serves the purpose of recognizing pedestrian attributes and holds significance in video surveillance applications, especially when close-up shots of faces and bodies are rarely accessible. It comprises a collection of 19,000 pedestrian images, encompassing 65 attributes, which consist of 61 binary attributes and 4 multi-class attributes. These images collectively depict 8,705 individuals.

\textbf{UPAR dataset~\cite{specker2023upar}.} UPAR is designed for pedestrian attribute recognition, person retrieval tasks in video surveillance and fashion retrieval. It is based on four well-known person attribute recognition datasets: PA100K, PETA, RAP v2~\cite{li2019richly}, and Market1501~\cite{lin2019improving}. UPAR unifies these datasets by providing 3.3 million additional annotations to harmonize 40 important binary attributes over 12 attribute categories across the datasets. 
Notably, we only use UPAR dataset to test generalization ability of model instead of using it to train. 

\textbf{People clothing segmentation (PCS) dataset~\cite{yang2014clothing,rajkumarlclothing}.} 
The dataset comprises of 1000 images with $825 \times 550$ pixels and  corresponding semantic segmentation masks. There are 59 semantic label classes, comprising of both the background and  clothing types.

\textbf{Metrics.}
Mean accuracy (mA) is employed as an evaluation metric for attribute recognition algorithms, which is formulated in Eq.~\ref{eq:ma}:
\begin{equation}
    mA=\frac{1}{2 N} \sum_{i=1}^{L}\left(\frac{T P_{i}}{P_{i}}+\frac{T N_{i}}{N_{i}}\right)
\label{eq:ma}    
\end{equation}
where $L$ is the number of attributes. $TP_i$ and $TN_i$ are the number of true positives and true negatives respectively, $P_i$ and $N_i$ are the number of all positive and negative examples respectively.

According to the existing methods, we utilize five metrics to evaluate model: mean accuracy (mA), accuracy (Accu), precision (Prec), recall (Rec), and F1 score (F1). 
\FloatBarrier
\subsection{Ablation study}
\label{sec:Ablation Study}
\begin{table}[]\small
\caption{The effect of AFSS (P1, P2, P3 represent approaches using the three fixed scale feature maps respectively. AFSS$^*$ and AFSS represent that AFSS select feature map for each attribute and each group of attributes respectively.)}
\centering
\scriptsize
\setlength{\tabcolsep}{1mm}{
\begin{tabular}{llllll}
\toprule
\textbf{Groups} & \textbf{P1}                      & \textbf{P2 }             & \textbf{P3}              & \textbf{AFSS$^*$}    & \textbf{AFSS} \\ \midrule
Head   & \textbf{80.09}          & 79.91           & 79.63           & 80.16    & 80.13          \\ 
Torso  & 87.71          & \textbf{88.13}           & 87.65           & 88.08    & 88.05          \\ 
Bottom & 80.11          & \textbf{80.42}          & 80.07           & 80.38    & 80.40           \\ 
All    & 80.39          & 80.45           & \textbf{80.75 }          & 80.77    & 80.71          \\ \bottomrule
\end{tabular}
}
\label{tab:AFSS role}
\end{table}

\textbf{Scale selection in AFSS.}
In order to find the optimal scale of feature map for each attribute group, we have conducted experiment on SSPNet-R (using PLE-R) with ResNet50 and mA results are shown in the middle columns of Tab.~\ref{tab:AFSS role} (P1, P2, P3).
It is clear to find that, the ``Head,'' ``Troso,'' ``Bottom,'' and ``All'' group obtains the best mA on feature map of P1, P2, P2, and P3 respectively.
Based on these observations, we think that attributes with small granularity require high resolution feature map, since it can provide detailed discriminative information to determine the existence and location of attribute. 
On the other hand, attributes with middle or large granularity prefer feature map P2 and P3 respectively, which is also in common sense for the design philosophy of feature pyramid network.

The right column of Tab.~\ref{tab:AFSS role} provides comparison between AFSS and AFSS$^*$. The AFSS aims to select feature scale for each group of attributes, while the AFSS$^*$ aims to select feature scale for each attribute. It is found that both of them can achieve similar performance, but AFSS is more generalizable and friendly in design when facing recognition tasks with a wide range of attributes. AFSS will keep as the default setting in the following experiments.

\textbf{Number of offset points in PLE-K.}
\begin{figure}\RawFloats
    \centering
    \begin{minipage}[t]{0.49\textwidth}
        \centering
        \includegraphics[width=1\textwidth]{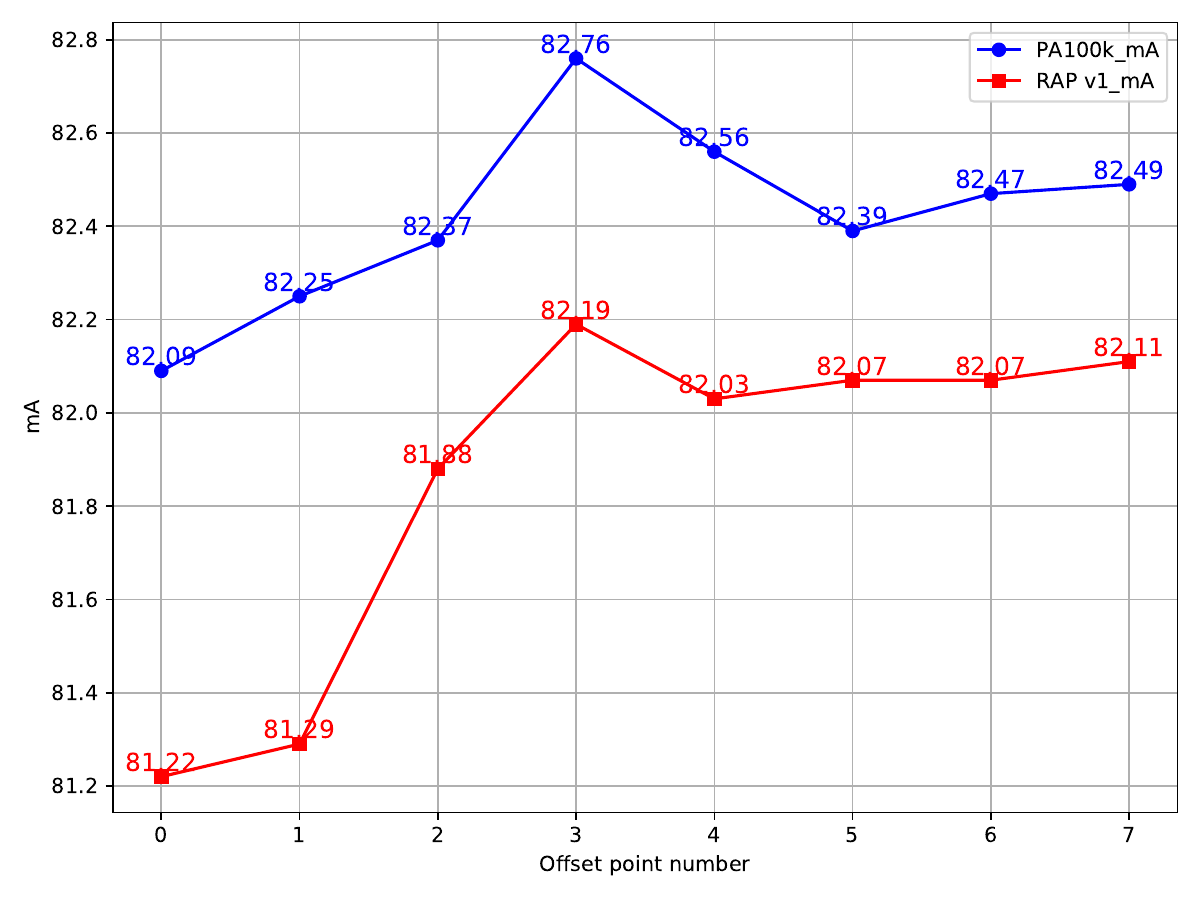}
        \caption{Performance with different number of offset points.}
        \label{fig:offset number}
    \end{minipage}
    \hfill
    \begin{minipage}[t]{0.49\textwidth}
        \centering
        \includegraphics[width=1\textwidth]{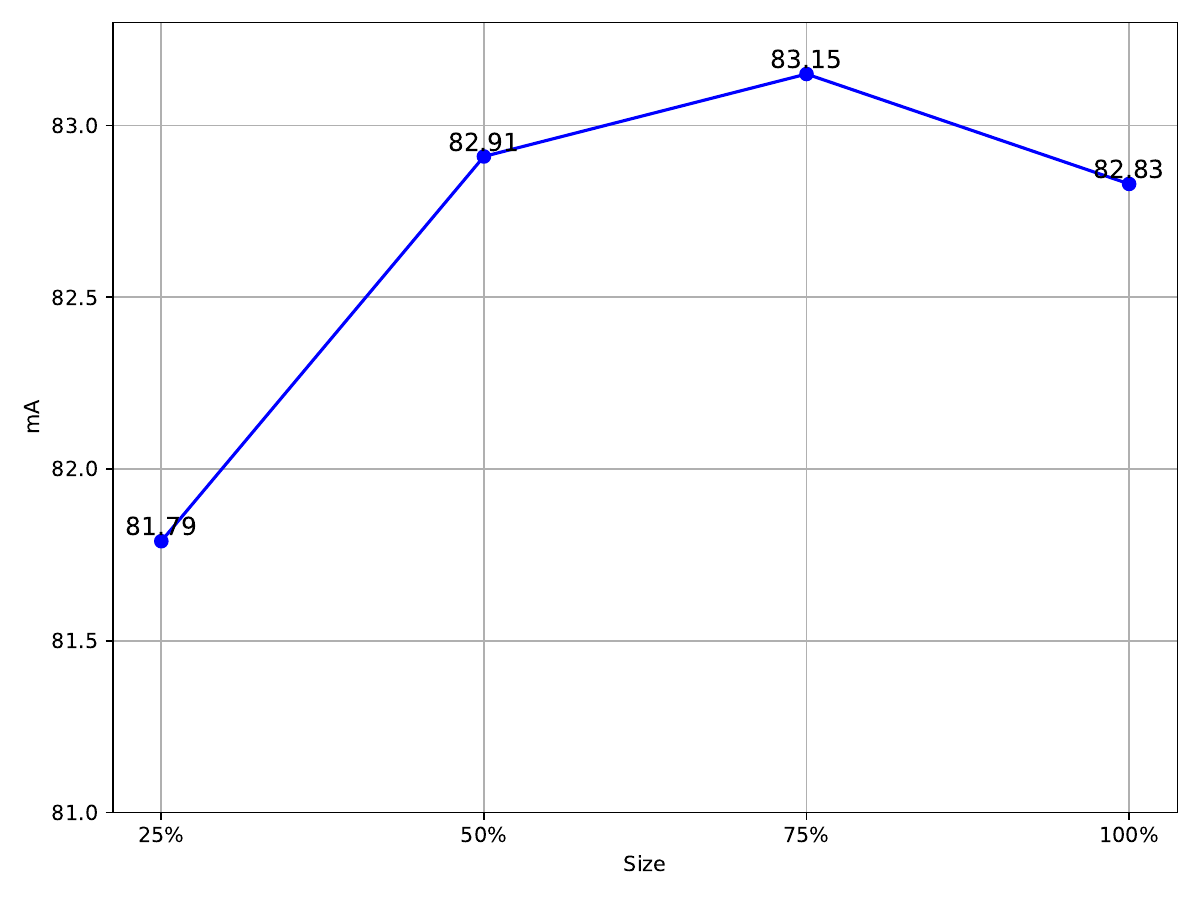}
        \caption{Performance with varying sparsity levels of sparse points.}
        \label{fig:sparsity}
    \end{minipage}
\end{figure}
Fig.~\ref{fig:offset number} illustrate the change curve of mA with different number of offset points on PLE-K with ResNet50. It is clear to find that the peak performance is reached when offset points number $M=3$, and we keep it as default value in the following experiments.

\textbf{Sampling ratios in PLE-S.} 
Fig.~\ref{fig:sparsity} provides the comparison of mA with different sampling ratio on PLE-S. It is found that taking 75\% of the feature points can achieve the best performance, comparing with sampling ratio of 25\%, 75\% and 100\%. 
Based on the observation, we infer that the sparsity of the sampled points has an impact on the performance of the model since less uniform sampling ratio ($\textless 50\%$) can lead to more discriminative information loss.
In the following experiments, we use 75$\%$ as the default sampling ratio.

\begin{table}[!t]\small
	\caption{The role of detail work (Offset: Offset points. Prior: Spatial prior knowledge. The symbols $\checkmark$ indicate that the corresponding component is included.)}
		\centering
        \scriptsize
		\begin{tabular}{lllll}
			\toprule
			\multirow{2}{*}{\textbf{Number}} &\multirow{2}{*}{\textbf{AFSS}}       & \multicolumn{2}{c}{\textbf{PLE-S}}   & \multirow{2}{*}{\textbf{mA}} \\ \cmidrule{3-4}
			&                   & \textbf{Offset}     & \textbf{Prior}    &                     \\ \midrule
			1&                       &            &          & 81.97               \\
 2& & \checkmark             &\checkmark           &82.81\\
			3&\checkmark             &            &          & 82.26               \\
			4&\checkmark   & \checkmark          &          & 82.60               \\
			5&\checkmark  &  & \checkmark         & 82.69      \\ 
			6&\checkmark  & \checkmark & \checkmark & \textbf{83.15}      \\ \bottomrule
		\end{tabular}
	\label{tab:ablations}
\end{table}

\textbf{Discussion on different components.}
Tab.~\ref{tab:ablations} provides detailed ablation experiments of AFSS and PLE modules. The first row is the baseline method, while the last row is our proposed method equipped with both AFSS and PLE modules. To make a fair and convincing evaluation, we deploy PLE based on sparse points as spatial prior knowledge in the Tab.~\ref{tab:ablations}.
By comparing the first and third row, AFSS can bring 0.29$\%$ improvements. When we use PLE-S, it can bring 0.89\% improvements from the third and sixth row. Similar results can also be obtained by comparing the first and second rows which shows that PLE-S can bring 0.74\% improvements. Within PLE-S, there are two components: offset and prior. When using offset alone, the learned offset points can improve performance by 0.34\% from the third and forth rows. When using prior alone, the spatial prior can provides a significant 0.43\% improvement from the third and fifth rows. Finally, by comparing the first and sixth row, AFSS+PLE-S can bring 1.18$\%$ additional improvements, which indicate the complementarity between two modules.

Compared to AFSS, PLE bring more performance improvement which indicates that feature extraction from precise attribute related locations is more important. 
Based on these observations, we can infer that both the scale prior (used in AFSS) and spatial prior (used in PLE) are effective in improving the performance of the model, and the model achieves the best performance when the two modules are used simultaneously.

\begin{table}[t]\small
\centering
\scriptsize
\caption{Comparing different backbone and efficiency on PA100k dataset.}
\begin{tabular}{llllllllll}
\toprule 
\textbf{Method }  & \textbf{Backbone}       & \textbf{mA}    & \textbf{Accu}  & \textbf{Pre}   & \textbf{Recall} & \textbf{F1}     & \textbf{Params}& \textbf{GFLOPs} &\textbf{FPS}\\ \midrule 
Basline  & ResNet50 & 81.97 & 80.20 & \textbf{88.06} & 88.17  & 88.11  & 23.56M& 2.69 &24\\
Basline  & Swin-S         & 82.19 & 80.35 & 87.85 & 88.51  & 88.18  & 48.86M& 7.74 &15\\ \midrule 
SSPNet-R & ResNet50 & 82.49 & 80.52 & 87.77 & 88.84  & 88.30  & 24.73M& 2.92&23\\
SSPNet-K & ResNet50 & 82.76 & 80.21 & 87.37 & 89.10  & 88.23  & 26.27M& 3.31&19\\
SSPNet-S & ResNet50 & 83.15 & 80.61 & 87.86 & 89.01  & 88.43  & 26.30M& 3.32&21\\ \midrule 
SSPNet-R & Swin-S          & 82.95 & 80.50 & 87.50 & 89.27  & 88.38  & 49.96M& 8.13&15\\
SSPNet-K & Swin-S          & 83.31 & 80.35 & 87.41 & \textbf{89.56}  & 88.47   & 52.43M& 8.65&11\\
SSPNet-S & Swin-S          & \textbf{83.58} & \textbf{80.63} & 87.79 & 89.32  & \textbf{88.55}  & 52.44M& 8.65&12\\ \bottomrule
\end{tabular}
\label{tab:backbone}
\end{table}

\textbf{Comparing different backbone and efficiency.} 
We have also conducted experiments by using  ResNet50 and Swin Transformer as backbones and efficiency comparison on the PA100k dataset which are shown in Tab.~\ref{tab:backbone}. We find that our SSPNet-S with Swin Transformer achieves the best performance in mA, Accu, Recall, and F1. 
Compared to the baseline method (ResNet50), our SSPNet-S (ResNet50) improves 1.18\%, 0.41\%, 0.01\%, and 0.25\% in mA, Accu, Recall, and F1, respectively. Compared to the baseline method (Swin Transformer), our SSPNet-S (Swin Transformer) improves 1.39\%, 0.28\%, 0.81\%, and 0.37\% in mA, Accu, Recall, and F1, respectively. Both SSPNet-R and SSPNet-K, which utilize ResNet50 and Swin Transformer as backbones, have demonstrated significant performance improvements compared to the baseline method. When using the same backbone, SSPNet-S outperforms SSPNet-K, and SSPNet-K outperforms SSPNet-R. Moreover, when utilizing the same PLE module, SSPNet with Swin-S achieves superior performance compared to SSPNet with ResNet50.

As shown in Tab.~\ref{tab:backbone}, compared to the baseline method, all SSPNet methods have few more parameters and a slightly decrease in FPS. Within the SSPNet family, SSPNet-R utilizes fewer parameters and computational resources in contrast to SSPNet-K and SSPNet-S, since it does not employ learnable offset points.

\begin{table}[t]\small
\centering
\caption{Comparison with state-of-the-art methods on PA100k dataset.}
\scriptsize
\resizebox{\linewidth}{!}{
\begin{tabular}{lllllll}
\toprule
\multirow{2}{*}{\textbf{Method}}  &\multirow{2}{*}{\textbf{Backbone}} &\multicolumn{5}{c}{\textbf{PA100k}}       \\ \cline{3-7}   
 & & \textbf{mA}             & \textbf{Accu}           & \textbf{Prec}           & \textbf{Recall}         & \textbf{F1 }            \\ \midrule
HP-Net~\cite{liu2017hydraplus} &InceptionNet &74.21 &72.19 &82.97 &82.09 &82.53 \\
PGDM~\cite{li2018pose} &CaffeNet &74.95 &73.08 &84.36 &82.24 &83.29  \\
VRKD~\cite{li2019pedestrian} &ResNet50     & 77.87          & 78.49          & 88.42         & 86.08          & 87.24     \\ 
VSGR~\cite{li2019visual}   &ResNet50     & 79.52          & 80.58          & \textbf{89.40}          & 87.15     & 88.26     \\ 
JLPLS-PAA~\cite{tan2019attention} &SE-BN-InceptionNet    & 81.61          & 78.89          & 86.83          & 87.73          & 87.27      \\ 
% DTM(Arxiv20)&ResNet50         & 81.63          & 77.57          & 84.27          & 89.02          & 86.58      \\ 
SSC~\cite{jia2021spatial}&ResNet50  & 81.87          & 78.89          & 85.98          & 89.10          & 86.87    \\
 CAS-SAL-FR~\cite{yang2021cascaded}& ResNet50& 82.86 & 79.64 & 86.81 & 88.78 &87.79 \\
 EALC~\cite{weng2023exploring}& EfficientNet-B4& 81.45& 80.27& 87.32& 88.98&88.14\\ \midrule
SSPNet (Ours) & ResNet50 & 83.15 & 80.61 & 87.86 & 89.01  & 88.43 \\ 
SSPNet (Ours) &Swin-S  &\textbf{83.58} & \textbf{80.63} & 87.79 & \textbf{89.32}  & \textbf{88.55}          \\ \bottomrule
\end{tabular}
}
\label{tab:pa100kres}
\end{table}

\begin{table}[!t]\small
\centering
\caption{Comparison with state-of-the-art methods on RAP v1 dataset. }
\scriptsize
\resizebox{\linewidth}{!}{
\begin{tabular}{lllllll}
\toprule
\multirow{2}{*}{\textbf{Method}} &\multirow{2}{*}{\textbf{Backbone}} &\multicolumn{5}{c}{\textbf{RAP v1}}       \\ \cline{3-7}
  & & \textbf{mA}  &\textbf{Accu} &\textbf{Prec} &\textbf{Recall} &\textbf{F1} \\ \midrule
HP-Net~\cite{liu2017hydraplus} &InceptionNet &76.12 &65.39 &77.33 &78.79 &78.05 \\
PGDM~\cite{li2018pose} &CaffeNet &74.31 &64.57 &78.86 &75.90 &77.35 \\
VRKD~\cite{li2019pedestrian} &ResNet50           & 78.30 & 69.79 & \textbf{82.13}& 80.35 & 81.23 \\ 
VSGR~\cite{li2019visual} &ResNet50           & 77.91 & 70.04 & 82.05 & 80.64 & 81.34 \\ 
JLPLS-PAA~\cite{tan2019attention} &SE-BN-InceptionNet            & 81.25                  & 67.91                    & 78.56                   & 81.45                      & 79.98                  \\ 
SSC~\cite{jia2021spatial}&ResNet50            & 82.77                & 68.37                    & 75.05                    & 87.49& 80.43\\
 PARFormer-L~\cite{fan2023parformer}& Transformer-based& \textbf{84.13}& 69.94& 79.63& \textbf{88.19}&81.35\\
 EALC~\cite{weng2023exploring}& EfficientNet-B4& 83.26& 69.65& 79.82& 83.61&81.67\\ \midrule 
SSPNet (Ours) & ResNet50 & 82.79 & 69.85 & 79.96 & 82.88  &81.39 \\ 
SSPNet (Ours) &Swin-S    & 83.24& \textbf{70.21}& 80.14           & 82.90            & \textbf{81.50}     \\ \bottomrule
\end{tabular}
}
\label{tab:rapv1res}
\end{table}

\begin{table}[!t]\small
\centering
\caption{Comparison with state-of-the-art methods on PETA dataset.}
\scriptsize
\resizebox{\linewidth}{!}{
\begin{tabular}{lllllll}
\toprule
\multirow{2}{*}{\textbf{Method}} &\multirow{2}{*}{\textbf{Backbone}} &\multicolumn{5}{c}{\textbf{PETA}}       \\ \cline{3-7}
  & & \textbf{mA}  &\textbf{Accu} &\textbf{Prec} &\textbf{Recall} &\textbf{F1} \\ \midrule
HP-Net~\cite{liu2017hydraplus} &InceptionNet &81.77 &76.13 &84.92 &83.24 &84.07 \\
PGDM~\cite{li2018pose} &CaffeNet &82.97 &78.08 &86.86 &84.68 &85.76 \\
VRKD~\cite{li2019pedestrian} &ResNet50           & 84.90 & 80.95 & 88.37 & 87.47 & 87.91 \\ 
VSGR~\cite{li2019visual} &ResNet50           & 85.21 & 81.82 & 88.43 & 88.42 & 88.42 \\ 
JLPLS-PAA~\cite{tan2019attention} &SE-BN-InceptionNet            & 84.88 & 79.46 & 87.42 & 86.33 & 86.87 \\              
SSC~\cite{jia2021spatial}&ResNet50            & 86.52 & 78.95 & 86.02& 87.12& 86.99\\
CAS-SAL-FR~\cite{yang2021cascaded}& ResNet50            & 86.40 & 79.93 & 87.03 & 87.33 &87.18 \\
EALC~\cite{weng2023exploring}& EfficientNet-B4& 86.84& 81.71& \textbf{88.58}& 88.23&88.40\\
DAFL~\cite{jia2022learning}& ResNet50            & 87.07 & -& -& -&86.40 \\ \midrule 
SSPNet (Ours) & ResNet50 & 88.14 & 81.96 & 88.53 & 88.27 &88.40 \\ 
SSPNet (Ours) &Swin-S    & \textbf{88.73}& \textbf{82.80}& 88.48 & \textbf{90.55}& \textbf{89.50}\\ \bottomrule
\end{tabular}
}
\label{tab:petares}
\end{table}

\subsection{Comparison with state-of-the-arts}
To validate the effectiveness of the proposed method, we use two public datasets and five evaluation metrics for comparative evaluation with SOTA PAR methods (HPNet~\cite{liu2017hydraplus}, PGDM~\cite{li2018pose}, VRKD~\cite{li2019pedestrian}, VSGR~\cite{li2019visual}, JLPLS-PAA~\cite{tan2019attention}, SSC~\cite{jia2021spatial}, CAS-SAL-FR~\cite{yang2021cascaded}, PARFormer-L~\cite{fan2023parformer}, EALC~\cite{weng2023exploring}, DAFL~\cite{jia2022learning})
on the PA100k, RAP v1, and PETA datasets. In this section, we utilize the best-performing SSPNet-S as a representative of SSPNet for comparison.

\textbf{Results on PA100k.} As shown in Tab.~\ref{tab:pa100kres}, our SSPNet with Swin-S ranks the first in terms of mA, Accu, Recall and F1 score, and our SSPNet with ResNet50 ranks the second with Prec. 
Compared to SSC, which establishes a spatial and semantic consistency framework to leverage inter-image relations and incorporate human priors, our proposed SSPNet with Swin-S outperforms it with improvements of 1.71\%, 1.74\%, 1.81\%, 0.22\%, and 1.68\% across all five metrics. When compared to CAS-SAL-FR, SSPNet with Swin-S respectively demonstrates performance enhancements of 0.72\%, 0.99\%, 0.98\%, 0.54\%, and 0.76\% across the same five metrics.

\textbf{Results on RAP v1.} As presented in Tab.~\ref{tab:rapv1res}, our SSPNet with Swin-S attains the highest Accu and F1 scores compared to other state-of-the-art approaches. In comparison to SSC, our proposed SSPNet with Swin-S exhibits notable improvements of 0.47\%, 1.84\%, 5.09\%, and 1.07\% for mA, Accu, Prec, and F1, respectively. Furthermore, our model outperforms PARFormer-L and EALC in terms of Accu, Prec, and F1. Additionally, SSPNet using ResNet50 also attained competitive results.

\textbf{Results on PETA}: As shown in Tab.~\ref{tab:petares}, SSPNet with both ResNet50 and Swin-S can achieve the SOTA results. SSPNet with Swin-S obtains the best performance in terms of mA, Accu, Recall, and F1. Compared to DAFL, it exhibits an improvement of 1.66\% in mA. Additionally, when compared to EALC, SSPNet achieves an increase of 1.10\% in F1. It is clear to observe that our approach consistly obtains the best result on the PETA dataset.
It is worth to mention that our SSPNet utilizes less information in the feature maps for using prior knowledge, compared to the baseline method. 
However, the learned fewer features preserve useful information because the model can focus more on the areas closely related to the attributes.

\begin{table}[!t]\small
\centering
\caption{Cross-dataset model test results(PA100k $\rightarrow$ RAP v1)} 
\scriptsize
\begin{tabular}{lllll}
\toprule
\textbf{Attributes}   & \textbf{Baseline} & \textbf{SSPNet-R}        & \textbf{SSPNet-K}  & \textbf{SSPNet-S}  \\ \midrule
Boots       & 50.12     & 51.63 (+1.51)          & 53.23 (+3.11)   & \textbf{53.61 (+3.49)}\\
Hat         & 63.82     & 65.33 (+1.51)          & 65.21 (+1.39)   & \textbf{65.78 (+1.96)}\\
HandBag     & 74.68     & 76.56 (+1.88)          & 77.21 (+2.53)   & \textbf{77.83 (+3.15)}\\
Glasses     & 78.52     & 79.12 (+0.60)          & 79.97 (+1.45)   & \textbf{80.87 (+2.35)}\\
Backpack    & 84.73     & 85.87 (+1.14)          & 86.24 (+1.51)   & \textbf{87.02(+2.29)}\\
ShortSleeve & 91.39     & 91.48  (+0.09)         & 91.41 (+0.02)   & \textbf{91.53 (+0.14)}\\
Female      & 90.20     & 90.95 (+0.75)          & 91.26 (+1.06)   & \textbf{91.31 (+1.11)}\\ \midrule
Average     & 76.21     & 77.28 (+1.07)          & 77.79 (+1.58)  &\textbf{78.28 (+2.07)} \\ \bottomrule
\end{tabular}
\label{tab:cross P-R}
\end{table}

\begin{table}[!t]\small
\centering
\caption{Cross-dataset model test results(RAP v1 $\rightarrow$ PA100k)} 
\scriptsize

\begin{tabular}{lllll}
\toprule
\textbf{Attributes}   & \textbf{Baseline} & \textbf{SSPNet-R}        & \textbf{SSPNet-K}  & \textbf{SSPNet-S}  \\ \midrule
Boots       & 80.52     & 82.18 (+1.66)          & 83.01 (+2.49)   & \textbf{83.24 (+2.72)}\\
Hat         & 68.54     & 70.18 (+1.64)          & 71.26 (+2.72)   & \textbf{71.31 (+2.77)}\\
HandBag     & 72.69     & 73.77 (+1.08)          & 74.63 (+1.94)   & \textbf{75.28 (+2.59)}\\
Glasses     & 75.81     & 76.31 (+0.50)          & 77.28 (+1.47)   & \textbf{78.49 (+2.68)}\\
Backpack    & 82.78     & 82.96 (+0.18)          & 83.84 (+1.06)   & \textbf{84.13 (+1.35)}\\
ShortSleeve & 76.37     & 77.12 (+0.75)          & 77.96 (+1.59)   & \textbf{78.47 (+2.10)}\\
Female      & 89.15     & 89.59 (+0.44)          & 89.42 (+0.27)   & \textbf{89.81 (+0.66)}\\ \midrule
Average     & 77.98     & 78.87 (+0.89)          & 79.63 (+1.65)   &\textbf{80.10 (+2.12)} \\ \bottomrule
\end{tabular}

\label{tab:cross R-P}
\end{table}

\begin{table}[]\small
\centering
\caption{Cross-dataset model test results(RAP v1 $\rightarrow$ UPAR datasets)}
\scriptsize
\begin{tabular}{lllll}
\toprule
\textbf{Attribute}   & \textbf{Baseline}   & \textbf{SSPNet-R}        & \textbf{SSPNet-K}  & \textbf{SSPNet-S}  \\ \midrule
BaldHead      & 69.71  & 70.69 (+0.98)          & 71.08 (+1.37)    &\textbf{71.73 (+2.02)}   \\
Glasses       & 78.65  & 80.21 (+1.56)          & 80.58 (+1.93)    &\textbf{81.37 (+2.72)}  \\
Hat           & 79.81  & 81.07 (+1.26)          & \textbf{82.49 (+2.68)}    &82.21 (+2.40)  \\
Backpack      & 82.64  & 83.59 (+0.95)          & 83.60 (+0.96)    &\textbf{83.82 (+1.18)}   \\
LongHair      & 86.88  & 88.29 (+1.41)          & 88.17 (+1.29)    &\textbf{89.37 (+2.49)}  \\
Female        & 95.02  & 95.97 (+0.95)          & 96.18 (+1.16)    &\textbf{97.22 (+2.20)}   \\ \midrule
Average       & 82.12  & 83.30 (+1.19)          & 83.68 (+1.57)    &\textbf{84.29 (+2.17)} \\ \bottomrule
\end{tabular}
\label{tab:cross R-U}
\end{table}

\subsection{Generalizability}
\label{Generalizability}
To evaluate the generalization capability of the proposed method, we have conducted three cross-dataset experiments. The mA of these shared attributes on the test datasets are reported separately.

\textbf{PA100k $\rightarrow$ RAP v1 dataset.} As shown in Tab.~\ref{tab:cross P-R}, the PAR results are tested on RAP v1 using the model trained on PA100k with a total of 7 attributes in common. 
We observe that the SSPNet-S gets the highest mA on all shared attributes and also the average mA among all methods. The SSPNet-R and SSPNet-K also get significant improvements compared to baseline method. The biggest boost comes when the SSPNet-S model recognizes the ``Boots'', which we believe is a gain from the combination of the prior region and the offset points. 

\textbf{RAP v1 $\rightarrow$ PA100k dataset.} As shown in Tab.~\ref{tab:cross R-P}, we use the PA100k to test model which is trained on the RAP v1. 
The SSPNet-S also gets the highest mA on all shared attributes and the highest average mA among all methods, which is similar to results of PA100k $\rightarrow$ RAP v1 cross-dataset test. The SSPNet-R, SSPNet-K and SSPNet-S get 0.89\%, 1.65\% and 2.12\% improvements on average mA of shared attributes, respectively. The largest boost comes from ``Hat'' and the second largest from ``Boots,'' both of which are fine-grained attributes and exhibit strong spatial priors. This indicates that our model excels at fine-grained attributes with strong spatial priors.

\textbf{RAP v1 $\rightarrow$ UPAR dataset.} 
We have conducted additional cross-dataset experiments on the UPAR dataset with the model trained on the RAP v1 dataset. To avoid data leaks, we exclude all the data from RAP dataset in the original UPAR dataset.
As shown in Tab.~\ref{tab:cross R-U}, SSPNet-S shows the best results except for the ``Hat'' attribute. Compared to baseline method, average mA of SSPNet-S gets 2.17$\%$ improvements which is consistent with the results of the above.
So we can infer that all SSPNet families exhibit better generalization performance by using prior knowledge, which demonstrates the potential for real-world applications.

To sum up, we find that our method favors fine-grained attributes with strong spatial prior, such as ``Boots,'' ``LongHair,'' and ``Hat''. While for coarse-grained attributes, our method can also obtain similar or higher performance compared to the baseline method on cross-datasets tests.

\begin{figure}[t]
	\centering
	\vspace{-0.15in}
	\begin{minipage}{1\linewidth}{
		\subfloat[Hat]{
			\includegraphics[width=0.49\linewidth,height=0.8in]{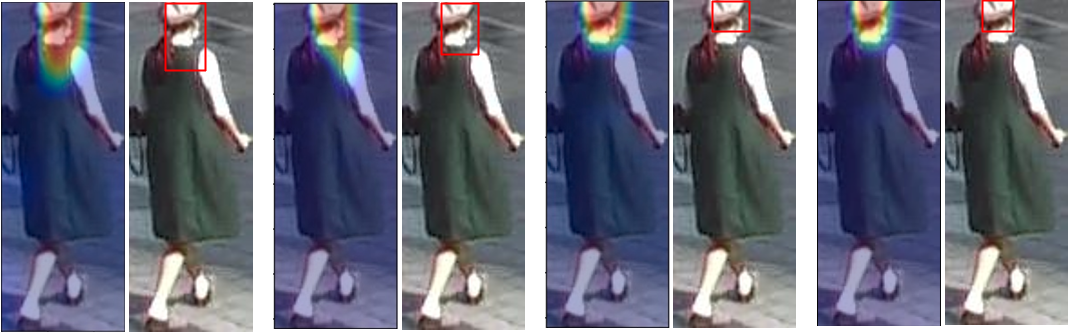}	   
                \label{fig:vis hat}		
  }\noindent
		\subfloat[Boots]{
		
			\includegraphics[width=0.49\linewidth,height=0.8in]{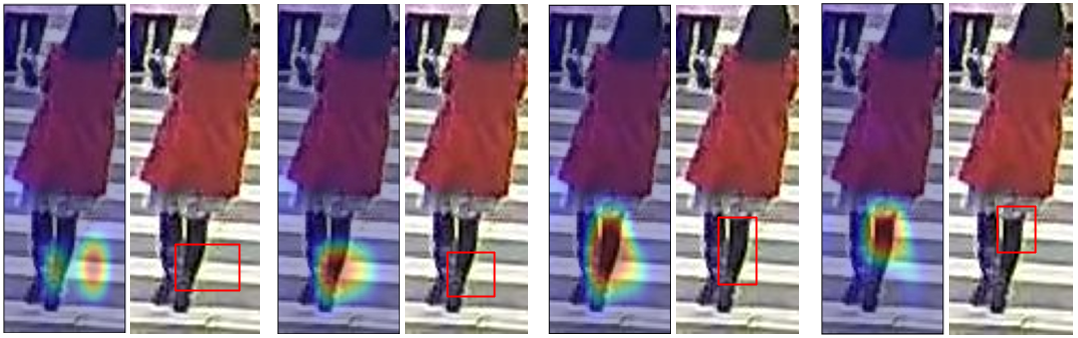}
                \label{fig:vis boots}		
  }
  }
	\end{minipage}
	\begin{minipage} {1\linewidth }{
		\subfloat[Shorts]{
			
			\includegraphics[width=0.49\linewidth,height=0.8in]{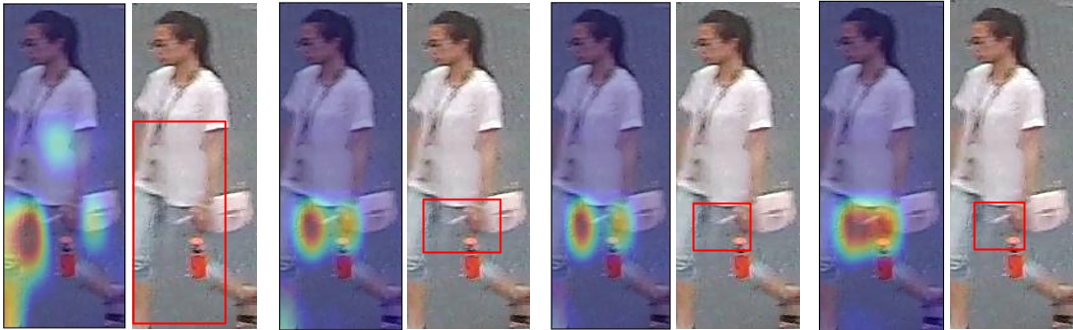}
                \label{fig:vis shorts}
			}
		\noindent
		\subfloat[Trousers]{
			
			\includegraphics[width=0.49\linewidth,height=0.8in]{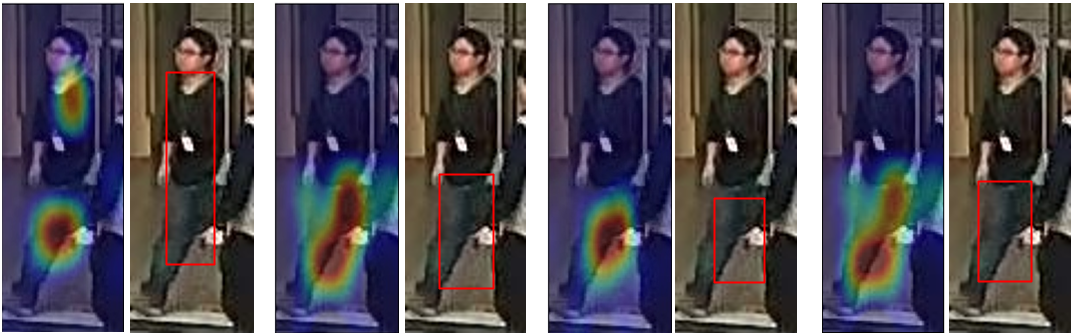}
                \label{fig:vis trousers}
			}
		}
	\end{minipage}
	\caption{Qualitative analysis of weakly supervised attribute localization(From left to right for each set of figures indicate baseline method, SSPNet-R, SSPNet-K, and SSPNet-S respectively. The red rectangular boxes are the attributes bounding boxed obtained from our method)}
	\label{fig:qualitative analysis}
\end{figure}

\subsection{Weakly supervised attribute localization}
\label{Weakly Supervised Attribute Localization}
\textbf{Qualitative analysis.} Fig.~\ref{fig:qualitative analysis} displays the heatmaps and weakly supervised attribute localization results generated using Grad-CAM-P. It can be seen that the attribute localization with baseline method is not satisfactory, with a phenomena of localization to irrelevant regions and insufficient fine localization. However, our methods that leverage prior knowledge can improve the localization of pedestrian attributes. 
After incorporating such prior knowledge, weakly supervised attribute localization results become more precise. 
Particularly, the model encounters a minor occlusion issue while attempting to localize ``Trousers"  in Fig.~\ref{fig:vis trousers}. Compared to the baseline method, SSPNet is able to locate the attribute more accurately.

The SSPNet-R can remove irrelevant regions because of prior regions, so we can see that heatmaps and bounding boxes more focus on attribute positions. SSPNet-K can further narrow the prior region by using more accurate human keypoints.
Besides, heatmaps and bounding boxes generated from SSPNet-S are also more accuracy due to the sparse sampling points compared to baseline method.

In summary, all the SSPNet families can produce different results for weakly supervised attribute localization, which demonstrates that the effectiveness of using spatial prior knowledge in different ways. 

\begin{figure}[t]
	\centering
	\vspace{-0.15in}
	\begin{minipage}{1\linewidth}
		\subfloat[Hat]{
		
			\includegraphics[width=1\linewidth,height=0.32\linewidth]{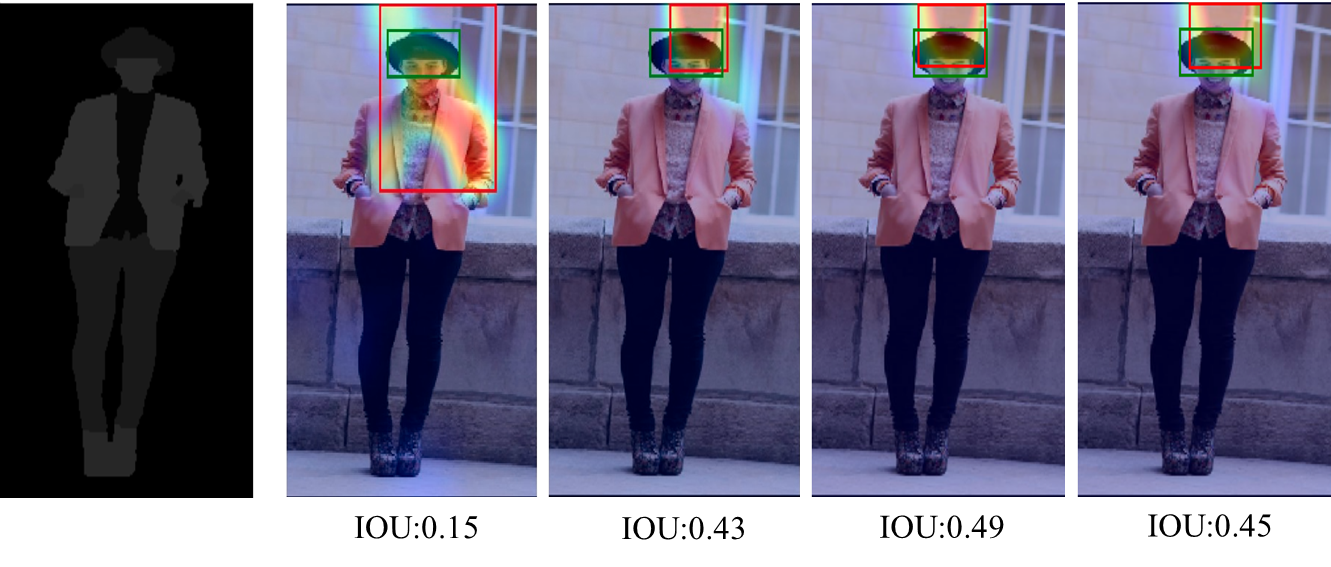}
   \label{fig:hat_iou}	
		}
	\end{minipage}
	\begin{minipage} {1\linewidth }{
		\subfloat[Boots]{
			
			\includegraphics[width=0.49\linewidth,height=1in]{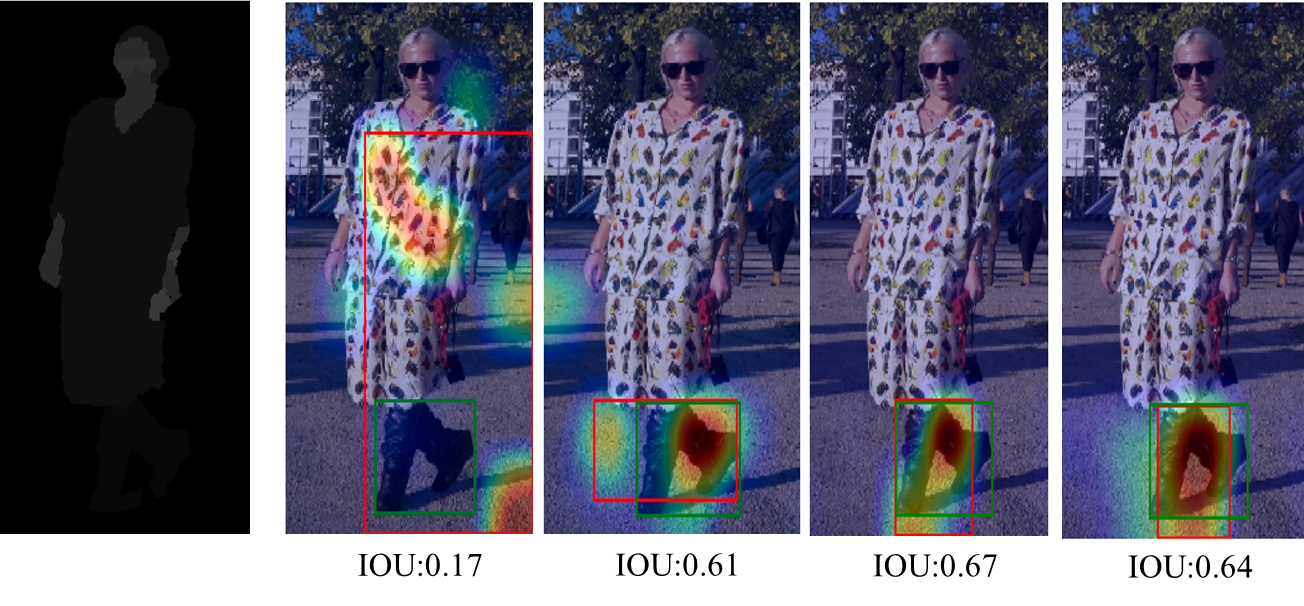}
   \label{fig:boots_iou}
			}
		\noindent
		\subfloat[Skirt]{
			
			\includegraphics[width=0.49\linewidth,height=1in]{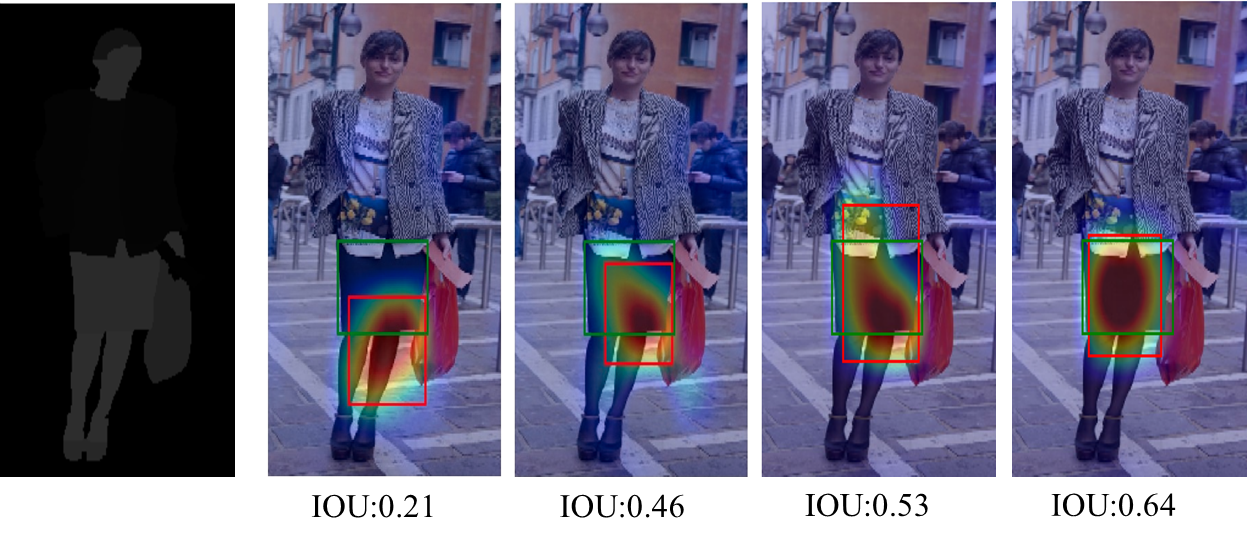}
   \label{fig:skirt_iou}
			}
		}
	\end{minipage}
	\caption{Quantitative analysis of weakly supervised attribute localization. (From left to right, for each image, it indicates the original segmentation annotation, baseline method, SSPNet-R, SSPNet-K, and SSPNet-S, respectively. The red and green rectangular boxes are the attribute bounding boxes obtained from our method and the ground truth, respectively.)}
	\label{fig:quantitative analysis}
\end{figure}

\textbf{Quantitative analysis.} To further evaluate the performance of our method on weakly supervised attribute localization quantitatively, we utilize the attribute segmentation annotation of the PCS dataset to generate accurate bounding boxes for the attributes. 
Similar to object detection, these bounding boxes can be used to compute IoU values to evaluate the performance of weakly supervised attribute localization.

As shown in Fig.~\ref{fig:quantitative analysis}, the IoU values are shown at the bottom of each image. For attribute ``Hat'' and ``Boots,'' the bounding boxes from the baseline method cover a large area of regions due to the use of global features.
However, the other three methods based on the spatial prior knowledge make the model focus on the correct location and obtain higher IoU values.
 
Compared to the ``Hat" and ``Boots," the ``Skirt" takes up a large area of the image which is less difficult to locate. It seems that the baseline method localization results are influenced by the background near the legs, leading to a lower IoU value. However, All of our methods obtain higher IoU, because the prior region excludes some of the distracting regions in advance. So our methods based on spatial prior knowledge exhibit superior weakly supervised attribute localization capabilities. Examples of our failure cases can be found in the appendix.

\begin{table}[t]\small
\caption{Quantitative result. (We computed the mean IoU, and the Pearson correlation coefficient (PCCs) about IoU and prediction confidence within the same attribute. B, R, K, and S represent baseline method, SSPNet-R, SSPNet-K, and SSPNet-S, respectively.)}
\scriptsize
\begin{tabular}{@{}lllllllll@{}}
\toprule
\multirow{2}{*}{\textbf{Attributes}} &
  \multicolumn{4}{l}{\textbf{IoU}} &
  \multicolumn{4}{l}{\textbf{PCCs}} \\ \cline{2-5} \cline{6-9} 
 &
  \multicolumn{1}{l}{\textbf{B}} &
  \multicolumn{1}{l}{\textbf{R}} &
  \multicolumn{1}{l}{\textbf{K}} &
  \textbf{S} &
  \multicolumn{1}{l}{\textbf{B}} &
  \multicolumn{1}{l}{\textbf{R}} &
  \multicolumn{1}{l}{\textbf{K}} &
  \textbf{S} \\ \midrule
Hat &
  \multicolumn{1}{l}{0.06} &
  \multicolumn{1}{l}{0.13} &
  \multicolumn{1}{l}{\textbf{0.17}} &
  0.16 &
  \multicolumn{1}{l}{0.26} &
  \multicolumn{1}{l}{0.40} &
  \multicolumn{1}{l}{0.41} &
  \textbf{0.43} \\
Boots &
  \multicolumn{1}{l}{0.09} &
  \multicolumn{1}{l}{0.25} &
  \multicolumn{1}{l}{\textbf{0.30}} &
  0.28 &
  \multicolumn{1}{l}{0.08} &
  \multicolumn{1}{l}{0.26} &
  \multicolumn{1}{l}{0.21} &
  \textbf{0.27} \\
Glasses &
  \multicolumn{1}{l}{0.04} &
  \multicolumn{1}{l}{0.27} &
  \multicolumn{1}{l}{\textbf{0.34}} &
  0.28 &
  \multicolumn{1}{l}{0.13} &
  \multicolumn{1}{l}{0.23} &
  \multicolumn{1}{l}{\textbf{0.28}} &
  0.26 \\
Skirt &
  \multicolumn{1}{l}{0.26} &
  \multicolumn{1}{l}{0.34} &
  \multicolumn{1}{l}{0.47} &
  0.47 &
  \multicolumn{1}{l}{0.25} &
  \multicolumn{1}{l}{0.37} &
  \multicolumn{1}{l}{\textbf{0.42}} &
  0.39 \\
Jeans &
  \multicolumn{1}{l}{0.30} &
  \multicolumn{1}{l}{0.37} &
  \multicolumn{1}{l}{0.39} &
  0.39 &
  \multicolumn{1}{l}{0.33} &
  \multicolumn{1}{l}{0.33} &
  \multicolumn{1}{l}{0.36} &
  \textbf{0.38} \\
Dress &
  \multicolumn{1}{l}{0.25} &
  \multicolumn{1}{l}{0.27} &
  \multicolumn{1}{l}{0.29} &
  \textbf{0.30} &
  \multicolumn{1}{l}{0.14} &
  \multicolumn{1}{l}{0.18} &
  \multicolumn{1}{l}{0.18} &
  0.18 \\
Shorts &
  \multicolumn{1}{l}{0.15} &
  \multicolumn{1}{l}{0.19} &
  \multicolumn{1}{l}{0.24} &
  \textbf{0.26} &
  \multicolumn{1}{l}{0.15} &
  \multicolumn{1}{l}{0.52} &
  \multicolumn{1}{l}{0.55} &
  \textbf{0.60} \\
Sunglasses &
  \multicolumn{1}{l}{0.04} &
  \multicolumn{1}{l}{0.06} &
  \multicolumn{1}{l}{0.12} &
  0.12 &
  \multicolumn{1}{l}{0.10} &
  \multicolumn{1}{l}{0.39} &
  \multicolumn{1}{l}{0.41} &
  \textbf{0.43} \\
T-shirt &
  \multicolumn{1}{l}{0.14} &
  \multicolumn{1}{l}{0.15} &
  \multicolumn{1}{l}{0.20} &
  \textbf{0.24} &
  \multicolumn{1}{l}{0.43} &
  \multicolumn{1}{l}{0.48} &
  \multicolumn{1}{l}{0.51} &
  \textbf{0.52} \\  \midrule 
Average &
  \multicolumn{1}{l}{0.15} &
  \multicolumn{1}{l}{0.23} &
  \multicolumn{1}{l}{0.28} &
  0.28 &
  \multicolumn{1}{l}{0.21} &
  \multicolumn{1}{l}{0.35} &
  \multicolumn{1}{l}{0.37} &
  \textbf{0.38} \\ \bottomrule
\end{tabular}
\label{tab:iou}
\end{table}
The comparison of average IoU results on the PCS dataset are shown in Tab.~\ref{tab:iou} (column ``IoU''). We have observed that SSPNet-K and SSPNet-S achieves the highest IoU for attribute localization, which is better than that of SSPNet-R. This can be well explained by the fact that SSPNet-K and SSPNet-S utilize more accurate spatial information compared to the whole prior region. 
Localizing fine-grained attributes, such as ``Hat,'' ``Boots,'' and ``Glasses,'' is a more challenging task and the baseline method yields poor results. However, SSPNet utilizing spatial priors leads to significant improvements in localization accuracy. 
Besides, localizing coarse-grained attributes, such as Skirt, SSPNet still shows significant performance improvements when compared to the baseline method.

It is worth noting that SSPNet-K performs better in localizing fine-grained attributes, while SSPNet-S performs better in localizing coarse-grained attributes from Tab.~\ref{tab:iou} (``K'' and ``S'' of column ``IoU''). This is because SSPNet-K benefits from the spatial prior of human keypoints, which provides more useful information for fine-grained attribute localization. However, when it comes to coarse-grained attribute localization, the limited number of human keypoints does not provide sufficient feature information compared with sparse points.

\textbf{Correlation analysis.} The comparison of the Pearson product-moment correlation coefficient (PCCs) 	\footnote{The PCCs is a numerical indicator that measures the strength of the linear relationship between two continuous variables, with values that range from -1 to 1.} about each sample IoU and attribute prediction confidence within the same attribute are shown in Tab.~\ref{tab:iou} (column ``PCCs''). 
All PCCs in our study were found to be positive, indicating a positive correlation between IoU values and attribute recognition confidence.

From Tab.~\ref{tab:iou} (column ``PCCs''), we can observe that SSPNet exhibits a significantly higher correlation compared to the baseline method. This indicates that SSPNet establishes a stronger positive correlation between attribute localization and attribute recognition confidence, thanks to the utilization of priors. Our proposed SSPNet approach effectively integrates attribute localization, prior knowledge, and attribute recognition confidence, resulting in a closer association. Notably, SSPNet-S achieves the highest average PCCs.

Furthermore, higher PCCs indicate a stronger positive correlation between localization IoU values and attribute recognition confidence. Therefore, improving attribute localization accuracy positively impacts attribute recognition.
This insight can motivates us to explore the possible way of using both attribute localization (IoU) and recognition (mA) as a new evaluation metric for PAR task.

\section{Conclusion}
We propose a novel PAR method guided by scale and spatial prior knowledge, to address the issues of interpretability for attribute locations and generalization performance for cross-dataset tests. 

Most relevant research overlooks the scale prior, and our experimental results suggest that choosing the appropriate feature map related to attribute granularity is beneficial to the PAR model. Additionally, it is noteworthy to mention that spatial prior knowledge can effectively guide the model to learn the proper attribute location and discriminative features, which improve the accuracy of PAR model and generalization ability. Furthermore, our hierarchical recognition framework and attribute grouping enable close-distance attributes to share features within the same group and prevent dependencies among different grouped attributes, thereby reducing the risk of over-fitting. Finally, we have also conducted a preliminary quantitative analysis of weakly supervised attribute localization. 
The experimental results on the public datasets validate the effectiveness of our proposed method in terms of recognition accuracy, generalization performance and location interpretability.

Our future work will try to extend our method to other attributes recognition tasks, such as face attributes, vehicle attributes and image retrieval. 

\section*{Acknowledgement}
This work was supported in part by NSF of China under Grant No. 61903164 and in part by NSF of Jiangsu Province in China under Grants BK20191427.

\section*{References}
\bibliography{mybibfile}
\clearpage
\appendix
\FloatBarrier
\section{Details of attribute group and spatial priors in PLE}

\begin{table}[h]\small
\renewcommand\arraystretch{0.8}
\scriptsize
\caption{Attributes grouping and Spatial prior region details. (The ``$\times$'' symbol is preceded by the range of height and followed by the width of the feature map.)}
\begin{tabular}{llll}
\toprule
\textbf{Input size} & \textbf{Groups} & \textbf{Attributes} & \textbf{Prior region} \\ \midrule
\multirow{4}{*}{\begin{tabular}[c]{@{}l@{}}P1: {[}0$\sim$63{]}$\times$48 \\ P2: {[}0$\sim$31{]}$\times$24 \\ P3: {[}0$\sim$15{]}$\times$12\end{tabular}} &
  Head &
  Hat, Glasses &
  \begin{tabular}[c]{@{}l@{}}P1: {[}0$\sim$24{]}$\times$48\\ P2: {[}0$\sim$12{]}$\times$24\\ P3: {[}0$\sim$6{]}$\times$12\end{tabular} \\ \cline{2-4} 
 &
  Torso &
  \begin{tabular}[c]{@{}l@{}}ShortSleeve, LongSleeve, UpperStride,\\ UpperLogo, UpperPlaid, UpperSplice,\\ Backpack\end{tabular} &
  \begin{tabular}[c]{@{}l@{}}P1: {[}8$\sim$40{]}$\times$48\\ P2: {[}4$\sim$20{]}$\times$24\\ P3: {[}2$\sim$10{]}$\times$12\end{tabular} \\ \cline{2-4} 
 &
  Bottom &
  \begin{tabular}[c]{@{}l@{}}LowerStripe, LowerPattern, Trousers,\\ Shorts, Boots\end{tabular} &
  \begin{tabular}[c]{@{}l@{}}P1: {[}32$\sim$63{]}$\times$48\\ P2: {[}16$\sim$31{]}$\times$24\\ P3: {[}8$\sim$15{]}$\times$12\end{tabular} \\ \cline{2-4} 
 &
  All &
  \begin{tabular}[c]{@{}l@{}}LongCoat, Skirt\&Dress, HandBag,\\ ShoulderBag,  HoldObjectsInFront, AgeOver60, \\Age18-60, AgeLess18, Female, Front, Side, Back\end{tabular} &
  \begin{tabular}[c]{@{}l@{}}P1: {[}0$\sim$63{]}$\times$48\\ P2: {[}0$\sim$31{]}$\times$24\\ P3: {[}0$\sim$15{]}$\times$12\end{tabular} \\ \bottomrule
\end{tabular}
\label{tab:attribute information}
\end{table}
We mention attribute groups in the AFSS module, and the detailed contents of the attribute groups are shown in the column ``Attributes'' of Tab.~\ref{tab:attribute information}. In the PLE-R module, we mention the height design of image blocks for each attribute group, and the specific data is displayed in the column ``Prior region'' of Tab.~\ref{tab:attribute information}.

\begin{figure}[t]
	\centering
	\captionsetup{aboveskip=0pt}
	\captionsetup{belowskip=0pt}
	\subfloat[Average image]{	
		\includegraphics[width=0.2\textwidth]{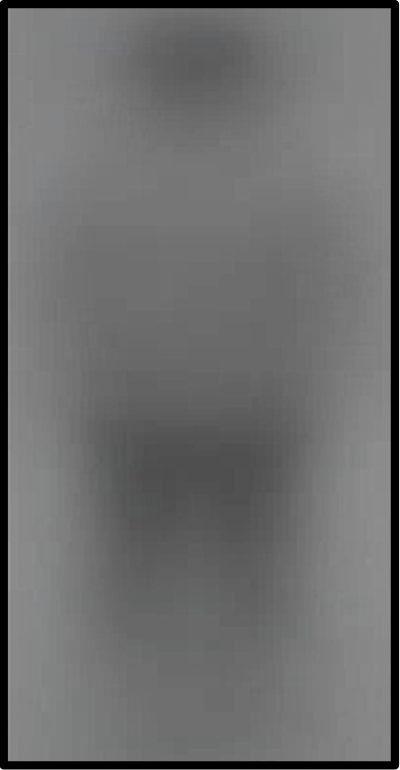}
		\label{fig:add image}
	}\hspace{8 mm}
	\subfloat[Points selection]{
		\includegraphics[width=0.23\textwidth]{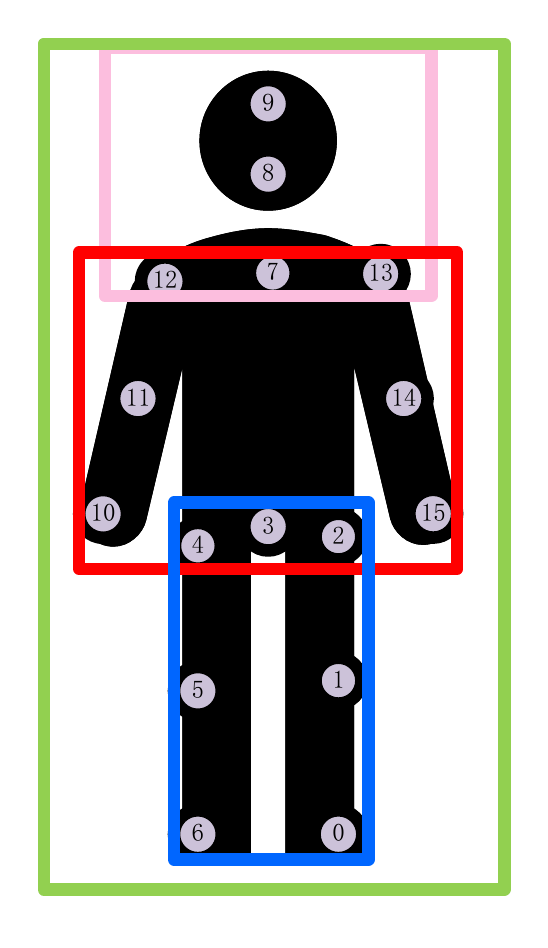}
		\label{fig:keypoint range}
	}
	\caption{Average image on PA100K dataset. (We overlay add all pedestrian images and average them to obtain the distribution of pedestrian bodies in the dataset.) and Points selection (It shows the selection of human keypoints through colored rectangular boxes for each attribute group, and details of point indexes through numbers in the points.)}
	\label{fig:people}
\end{figure}

Our inspiration for image block partitioning to obtain prior regions comes from the overlaid and averaged pedestrian images, as shown in Fig.~\ref{fig:add image}. These images exhibit clear outlines of the human body. 

As illustrated in Fig.~\ref{fig:keypoint range}, we present the 16 human keypoints along with their corresponding indexes in the PLE-K module. The keypoints surrounded by rectangles of different colors represent the reference points selected for the respective attribute groups. For detailed information regarding the selected human keypoints in each attribute group, please refer to Tab.~\ref{tab:keypoint information}.

\begin{table}[h]\small
\caption{Selected points of PLE-K. (Four attributes group of human parts and keypoints index)}
\begin{tabular}{ll}
\toprule
\textbf{Group}  &\textbf{Parts name and keypoints index}  \\ \midrule
head   &head(7$-$9), shoulder(12, 13)   \\ 
torso  &shoulder(12, 13), arms(10$-$15), hip(2, 3, 6)\\ 
bottom &hip(2, 3, 6), legs(0$-$5)\\ 
all    & whole body(0$-$15)\\ \bottomrule
\end{tabular}
\label{tab:keypoint information}
\end{table}

\FloatBarrier
\section{Case analysis}
\subsection{Discussion on hard cases}
\begin{figure}[h]
    \centering
    \includegraphics[width=1\textwidth]{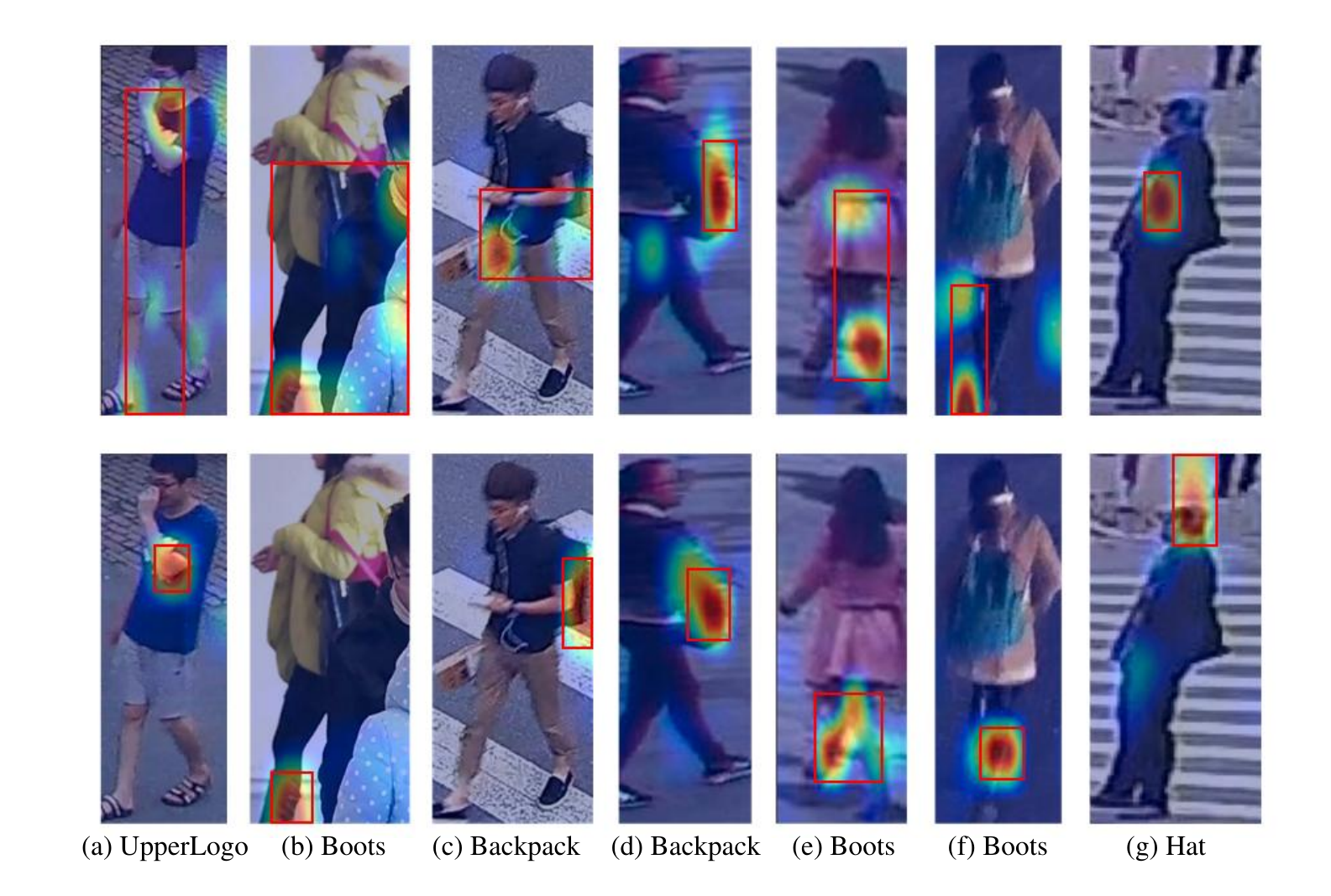}
    \caption{Hard cases. (The upper column presents the baseline results, while the lower column displays the results for SSPNet.)}
    \label{fig:hard case}
\end{figure}
Fig.~\ref{fig:hard case} presents results on hard cases. In (a), (b), and (c), attributes suffer from occlusion. In (b), (e), and (f), scale deformation occurs under the same attribute due to varying distances. (d) and (e) demonstrate deformation caused by pedestrian movement, and (g) showcases a fine-grained attribute. SSPNet consistently outperforms the baseline method by accurately localizing attributes while avoiding irrelevant regions. Our model excels in attribute localization, addressing issues related to occlusion, deformation, and fine-grained details.
\begin{figure}[t]
    \centering
    \includegraphics[width=0.75\textwidth, height=0.38\textwidth]{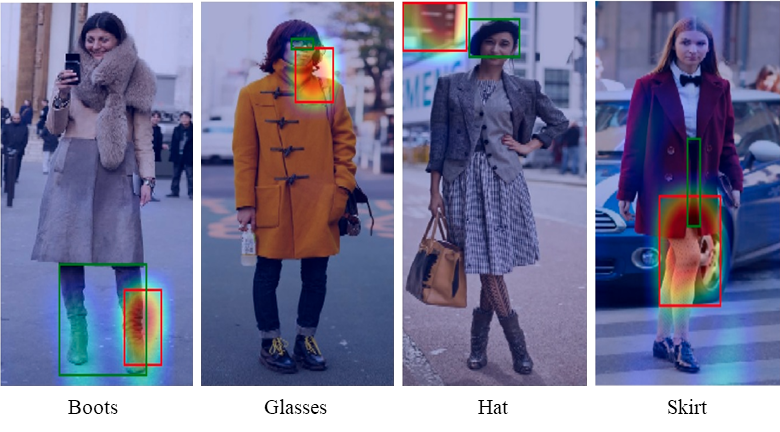}
    \caption{Fail cases. (These failed cases involve attribute labels that are predicted correctly, but the predicted locations significantly deviate from our expectations. Boots, Glasses, Hat, and Skirt visualization heatmaps are obtained from SSPNet-K, SSPNet-R, SSPNet-S, and SSPNet-R, respectively)}
    \label{fig:failed case}
\end{figure}
% \FloatBarrier
\subsection{Failure cases and limitation}
Fig.~\ref{fig:failed case} illustrates the failure cases and reflects the limitations of our method in weakly supervised attribute localization.
``Boots'' in Fig.~\ref{fig:failed case} indicates that when the attributes requiring localization are widely spread (e.g., two boots), our method may miss some positions. This is partly due to the weak supervision signal provided by the label level. ``Glasses'' in Fig.~\ref{fig:failed case} demonstrates the limitations of our method in localizing extremely fine-grained attribute positions. One possible reason is that even when AFSS selects the feature map at the maximum scale (downsampled by a factor of 4) for attribute recognition, it may still miss the fine-grained features of the attributes. ``Hat'' in Fig.~\ref{fig:failed case} indicates that SSPNet-S has room for improvement in terms of its ability to filter complex backgrounds that are similar to the attributes being recognized and localized. ``Skirt'' in Fig.~\ref{fig:failed case} demonstrates that the localization performance is not satisfactory when dealing with special samples where the attribute is heavily occluded.
\end{document}